\documentclass[lettersize,journal]{IEEEtran}
\usepackage{amsmath,amsfonts}
\usepackage{amsmath}
\usepackage{amssymb}
\usepackage{array}
\usepackage[caption=false,font=footnotesize,labelfont=sf,textfont=sf]{subfig}
\usepackage{textcomp}
\usepackage{stfloats}
\usepackage{url}
\usepackage{verbatim}
\usepackage{graphicx}
\usepackage{cite}
\newtheorem{definition}{Definition}

\usepackage{epstopdf}

\usepackage{multirow}

\usepackage[table,xcdraw]{xcolor}
\usepackage[ruled,vlined]{algorithm2e}

\begin{document}

\title{Towards Efficient Target-Level Machine Unlearning Based on Essential Graph}

\author{Heng Xu,~
        Tianqing Zhu*,~\IEEEmembership{Member,~IEEE,}
        Lefeng Zhang,~
        Wanlei~Zhou,~\IEEEmembership{Senior Member,~IEEE,}
        and~Wei~Zhao,~\IEEEmembership{Fellow,~IEEE}
\thanks{
*Tianqing Zhu is the corresponding author. 
Heng Xu is with the Centre for Cyber Security and Privacy and the School of Computer Science, University of Technology Sydney, Ultimo, NSW 2007, Australia (e-mail: heng.xu-2@student.uts.edu.au). Tianqing Zhu, Lefeng Zhang and Wanlei Zhou are with the City University of Macau, Macau (e-mail: tqzhu@cityu.edu.mo; lfzhang@cityu.edu.mo; wlzhou@cityu.edu.mo). Wei Zhao is with Shenzhen Institute of Advanced Technology, University of Chinese Academy of Sciences, Shenzhen, China (e-mail: zhao.wei@siat.ac.cn).
}
}

\markboth{Journal of \LaTeX\ Class Files,~Vol.~14, No.~8, August~2021}%
{Shell \MakeLowercase{\textit{et al.}}: A Sample Article Using IEEEtran.cls for IEEE Journals}


\maketitle

\begin{abstract}
Machine unlearning is an emerging technology that has come to attract widespread attention. A number of factors, including regulations and laws, privacy, and usability concerns, have resulted in this need to allow a trained model to forget some of its training data. Existing studies of machine unlearning mainly focus on unlearning requests that forget a cluster of instances or all instances from one class. While these approaches are effective in removing instances, they do not scale to scenarios where partial targets within an instance need to be forgotten. For example, one would like to only unlearn a person from all instances that simultaneously contain the person and other targets. Directly migrating instance-level unlearning to target-level unlearning will reduce the performance of the model after the unlearning process, or fail to erase information completely. To address these concerns, we have proposed a more effective and efficient unlearning scheme that focuses on removing partial targets from the model, which we name ``target unlearning". Specifically, we first construct an essential graph data structure to describe the relationships between all important parameters that are selected based on the model explanation method. After that, we simultaneously filter parameters that are also important for the remaining targets and use the pruning-based unlearning method, which is a simple but effective solution to remove information about the target that needs to be forgotten. Experiments with different training models on various datasets demonstrate the effectiveness of the proposed approach.
\end{abstract}

\begin{IEEEkeywords}
Machine unlearning, data privacy, model explanation, target unlearning, deep learning,
\end{IEEEkeywords}

\section{Introduction}
\label{sec:introduction}

\IEEEPARstart{M}{achine} unlearning refers to mechanisms that remove the influence of partial training instances from a machine learning model. It has recently emerged as a promising technology due to several factors, including regulations and laws, privacy, and model utility~\cite{DBLP:journals/csur/XuZZZY24,DBLP:conf/sp/CaoY15,DBLP:conf/icse/LiH0CL21}. Simply removing those instances from the dataset is not enough, as the impact of the removing information also exists in the trained model, and can be revealed by some machine learning attacks~\cite{DBLP:conf/sp/ShokriSSS17}. 
The most straightforward way for machine unlearning is to retrain the model from scratch after removing the instances that are requested to be unlearned from the training dataset. However, as machine learning typically now involves enormous training datasets, this straightforward method would lead to prohibitively expensive computational and time costs~\cite{DBLP:conf/sp/CaoY15}. To reduce the overhead, recent studies have proposed a number of different unlearning techniques~\cite{DBLP:conf/sp/BourtouleCCJTZL21,DBLP:conf/aaai/GravesNG21,DBLP:conf/nips/GuptaJNRSW21,DBLP:conf/icml/GuoGHM20}. 

\begin{figure}[!t]
    \centering
    \includegraphics[width=0.48\textwidth]{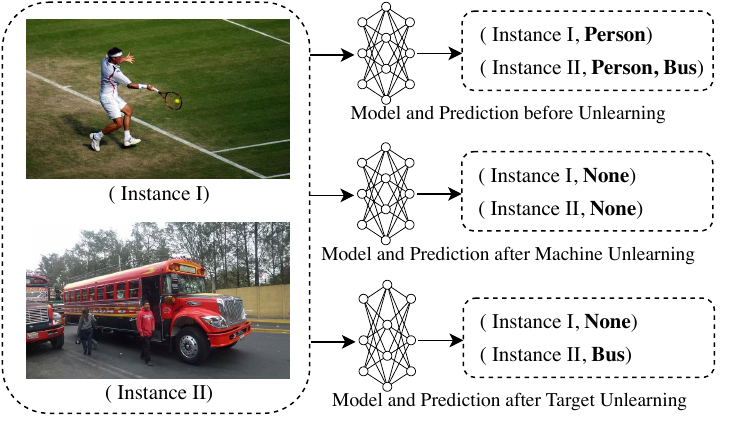}
    \caption{Target unlearning.}
    \label{fig:example}
\end{figure}

Current machine unlearning works mainly focus on unlearning partial instances selectively, such that they only remove information at instance level~\cite{DBLP:conf/ijcai/YanLG0L022}. We argue that we may need to support machine unlearning at the target level. We use \textit{target} to denote the object that appears in an instance, and the scheme is defined as \textit{target unlearning}. The requirements for unlearning targets can arise from eliminating the model's capabilities for unimportant, sensitive, or abnormal objects. For instance, it can involve removing a specific category from a multi-label classification task or eliminating a particular object from an object detection task. Figure~\ref{fig:example} shows the difference between traditional instance-level unlearning and target unlearning, where the two instances on the left side contain different numbers of targets. Instance I contains a person as one target, while instance II contains two objects, a person and a bus, so we consider them as two targets. When a request to remove information about the person is received, traditional instance-level unlearning schemes will directly remove two instances; however, the bus will also be affected as instance II will also be removed. For target unlearning, it will only remove accurate information about the person from each instance.

Therefore, how to remove the impact of targets accurately is our research question. There are mainly two unique challenges on this question. First, in the context of target unlearning, one instance may contain multiple targets, and those targets that need to be removed are tightly embedded in each instance, which makes it difficult to separate them in the instance. Moreover, multiple targets in each instance are interactive, and the machine learning model may further interact with diverse targets in different layers. it is difficult to remove the effect of a certain target while minimizing collateral damage to model performance for other remaining targets.

To address the above concerns, we propose an interpretive approach to finding parameters that have the greatest impact on targets that needs to be unlearned, and then pruning these parameters to erase information~\cite{DBLP:conf/cvpr/GolatkarAS20,DBLP:conf/icml/GuoGHM20,DBLP:conf/www/Wang0XQ22}. Specifically, we use model explanation method to find the most influential parameters for a particular region within an instance~\cite{DBLP:conf/cvpr/WangWDYZDMH20,DBLP:journals/tip/JiangZHCW21}. This approach can effectively separate various targets from the instance, and directly identify the model parameters that have a great impact on these separated targets. However, interpretive methods typically focus on a single layer and pruning all influential parameters within that layer may negatively impact the performance of the model for the remaining data. First, due to the specific structure of deep neural networks, considering parameter pruning in only one layer may not satisfy the effectiveness of unlearning. For example, focusing solely on the influential parameters in the last layer and then pruning those parameters to erase the information may not be sufficient for preserving privacy. Once an attacker has obtained all parameters of the model in a white-box scenario, they can still infer information from the penultimate level about the unlearned target. Second, the parameters selected based on the model explanation may also be important for other targets, and blindly pruning all those selected parameters will reduce the performance of the unlearned model.

To tackle those challenges, we construct an essential graph structure to describe the relationships of all influential parameters in each layer. This essential graph also provides a way to consider the performance of the unlearned model on remaining targets. It will simultaneously consider both the remaining target and the unlearning target to filter the parameters during the balanced process. To validate our proposed unlearning scheme, we evaluate multiple machine learning tasks under different models and various datasets, which demonstrates that our proposed scheme can effectively achieve the removal of target-level information and the ability to reasonably weigh the performance of the remaining targets. The main contributions of this paper can be summarized as follows\footnote{Our code can be found in: https://github.com/IMoonKeyBoy/Towards-Efficient-Target-Level-Machine-Unlearning-Based-on-Essential-Graph}:

\begin{itemize}%
    \item We initiate the exploration of machine unlearning at the target level by defining the problem of efficient target unlearning, including its formalization and challenges.
    \item We investigate target unlearning from a novel interpretative perspective and selectively prune the most influential parameters to achieve the unlearning purpose.
    \item We propose an essential graph data structure to describe the relationships between influential parameters of multiple layers and simultaneously consider the remaining data's impact to mitigate the performance degradation. 
    \item We validate our target unlearning scheme across various datasets, models, and machine learning tasks, demonstrating that our design can maintain model utility while achieving significant unlearning effectiveness.
\end{itemize}

\section{Related work}
\label{sec:relatedwork}

\subsection{Machine Unlearning}
The machine learning community has proposed numerous unlearning schemes in response to \textit{the right to be forgotten}. In our previously published survey paper~\cite{DBLP:journals/csur/XuZZZY24}, we conducted a comprehensive survey encompassing recent studies on machine unlearning. This survey thoroughly summarized several crucial facets, including: (i) the motivation behind machine unlearning; (ii) the objectives and desired outcomes associated with the unlearning process; (iii) a novel taxonomy for systematically categorizing existing machine unlearning schemes according to their rationale and strategies; and (iv) the characteristics as well as the advantages and disadvantages of existing verification approaches.

In general, existing unlearning works mainly focus on the following two types of unlearning requests: instance-level and class-level. Instance-level unlearning considers the unlearning request that removes information about one or a cluster of instances from the model. Cao et al.~\cite{DBLP:conf/sp/CaoY15} converted machine learning models into summation forms and used these summations to build statistical query learning models. The unlearning operation recomputed the affected summations for instances that needed to be unlearned. Bourtoule et al.~\cite{DBLP:conf/sp/BourtouleCCJTZL21} presented a "Sharded, Isolated, Sliced, and Aggregated" (SISA) architecture, similar to the existing distributed training methodologies, to achieve the unlearning purposes. Guo et al.~\cite{DBLP:conf/icml/GuoGHM20} proposed \textit{certified removal}, an unlearning scheme inspired by differential privacy~\cite{DBLP:journals/tkde/ZhuYWZY22}. It first limited the maximum difference between the unlearned and retrained models, then used model shifting to achieve unlearning. Golatkar et al.~\cite{DBLP:conf/cvpr/GolatkarARPS21} divided training data into \textit{core} and \textit{user} data. They trained the model non-convexly on \textit{core} data, and then on \textit{user} data with a quadratic loss function. This allows effective unlearning of \textit{user} data using existing quadratic unlearning schemes~\cite{DBLP:conf/icml/GuoGHM20}. Chundawat et al.~\cite{DBLP:conf/aaai/ChundawatTMK23} proposed a teacher-student framework using knowledge distillation, where both competent and incompetent teachers transfer knowledge. The incompetent teacher unlearns specific data, while the competent teacher retains general knowledge. Kurmanji et al.~\cite{DBLP:conf/nips/KurmanjiTHT23} achieved unlearning by constraining the model output using KL-divergence. For the remaining dataset, the output closely matches the original model's output, while for the unlearning dataset, it should be as dissimilar as possible. Foster et al.~\cite{DBLP:conf/aaai/FosterSB24} proposed Selective Synaptic Dampening (SSD) for unlearning. SSD used the Fisher Information Matrix (FIM) to identify and dampen parameters most influenced by the unlearning instance to achieve the unlearning purpose. Schelter et al.~\cite{DBLP:conf/sigmod/SchelterGD21}, and Brophy et al.~\cite{DBLP:conf/icml/BrophyL21} considered the effective unlearning methods in tree-based models, while Nguyen et al.~\cite{DBLP:conf/nips/NguyenLJ20} focused on Bayesian models.

Class-level unlearning means that the request is to remove information about all instances belonging to a particular class. Baumhauer et al.~\cite{DBLP:journals/ml/BaumhauerSZ22} transformed existing logit-based classifiers into an integrated model, decomposable into a feature extractor, followed by logistic regression. To unlearn the given class data, four filtration methods were proposed: naive unlearning, normalization, randomization, and zeroing. Graves et al.~\cite{DBLP:conf/aaai/GravesNG21} provided a framework based on random relabelling and retraining schemes. Each instance of the unlearning class was relabeled with randomly selected incorrect labels. Then, the machine learning model was retrained based on the modified dataset for some number of iterations to unlearn all those instances in one class. Wang et al.~\cite{DBLP:conf/www/Wang0XQ22} analyzed the problem of class-level unlearning in federated learning. They introduced the concept of the term frequency-inverse document frequency (TF-IDF) to quantify the class discrimination of the channels. An unlearning procedure via channel pruning~\cite{DBLP:conf/cvpr/GuoWLY20} was also provided, followed by a fine-tuning process to recover the performance of the pruned model. Ayush et al.~\cite{TNNLSAyush2023} introduced an unlearning method based on impairing and repairing. They first generated a noise matrix to target influential patterns of the unlearning class, then used it to fine-tune the model. After impairing, they further fine-tuned the model with the remaining data to maintain performance. For other types of unlearning requests, Chen et al.~\cite{DBLP:conf/ccs/Chen000H022} introduced the strategy established by Bourtoule et al.~\cite{DBLP:conf/sp/BourtouleCCJTZL21} to graph data, and provides two balanced graph partition algorithms with a learning-based aggregation method. Gupta et al.~\cite{DBLP:conf/nips/GuptaJNRSW21} proposed an unlearning method that focuses on sequential unlearning requests using a variant of the SISA framework as well as a differential privacy aggregation method. Alexander et al.~\cite{DBLP:conf/ndss/WarneckePWR23} used influence functions to formulate unlearning as a closed-form update to the model parameters, enabling efficient unlearning of feature or label information for tabular datasets.

Unfortunately, the reviewed unlearning works cannot handle target-level unlearning requests~\cite{DBLP:conf/nips/KurmanjiTHT23,TNNLSAyush2023,DBLP:conf/ndss/WarneckePWR23}. Target-level unlearning considers the unlearning of part of an instance, while instance- or class-level unlearning schemes focus on removing the information from whole instances. If these instance-level schemes are applied to the target-level unlearning, it will lead to two extreme results: over-unlearning or under-unlearning. First, if there are many instances containing the unlearning target, omitting all instances that contain the unlearning target will inevitably reduce the performance of the model after the unlearning process. This is because those instances may also contain other targets~(shown in Figure~\ref{fig:example}). Second, if only considering unlearning partial instances that only contain the unlearning target, the unlearned model will still contain the contribution of unlearning targets since there may be some instances that contain more than one target, one of which is unlearning target. Obviously, the process of unlearning is not limited to just eliminating the information from class or instance but should also include the unlearning request that focuses on a more fine-grained level: target-level.

\subsection{Model Explanation}
\label{subsec:modelexplanation}
In recent years, the interpretability of neural networks has received increased attention~\cite{DBLP:journals/csur/GuidottiMRTGP19,DBLP:journals/tnn/TjoaG21,DBLP:journals/tkde/LiCSBGQWGZXC22}. Different techniques have been developed to investigate the working mechanisms encoded inside the deep neural model. Class Activation Mapping~(CAM) is a well-known technique for generating heatmaps that correlate to the discriminative regions in one image for CNN-based models. Zhou et al. first proposed this technique and used the weighted linear combination of feature maps of the final convolutional layer to achieve the generation of CAM~\cite{DBLP:conf/cvpr/ZhouKLOT16}. In their schemes, the weights used for feature map combination were obtained from the trained fully-connected layer. This, however, limits this scheme's prowess to CNNs since only CNN models where the penultimate two layers are Global Average Pooling~(GAP) and linear fully-connected can be used, or it needs to retrain multiple additional linear classifiers after training the initial model. Grad-CAM was built to address the above issues~\cite{DBLP:conf/iccv/SelvarajuCDVPB17}. Grad-CAM utilizes the gradients backpropagating from the output node to compute the weights for each feature map. CAM and Grad-CAM provide a measurement for the importance of each feature map towards the overall decision of CNN. However, they do not consider pixel-level importance assessment, which can lead to performance drops when localizing multiple occurrences of the same class or localizing the entire object. Chattopadhyay et al.~\cite{DBLP:conf/wacv/ChattopadhyaySH18} proposed Grad-CAM++ and introduced pixel-level weighting. In gradient-free visual explanation methods, Wang et al.~\cite{DBLP:conf/cvpr/WangWDYZDMH20} proposed Score-CAM, which used the feature map as a mask and combined it with the original image to obtain its forward-passing score on the target class, those scores were then regarded as the combination weights.

The above schemes used different techniques to identify the discriminative regions, and the intermediate weights indicate how important each feature map for those regions is~\cite{DBLP:conf/cvpr/WangWDYZDMH20,DBLP:journals/tip/JiangZHCW21}. For target unlearning, the target in each image can also be regarded as a discriminative region. In this case, the above important relationship can be used to represent the influence of the corresponding feature map on the unlearning target. This varying importance of feature maps provides a feasible way for unlearning target data, i.e., by directly manipulating the corresponding model parameters to remove the information contained in those model parameters~\cite{DBLP:conf/cvpr/LiLZLDWHJ19,DBLP:conf/iclr/MorcosBRB18,DBLP:journals/corr/abs-1901-08644}. 

\begin{table}
  \caption{Notations}
  \renewcommand{\arraystretch}{1.2}
  \label{tab:notations}
  \centering
  \begin{tabular}{c|c}
    \hline
    Notations &  Explanation \\
    \hline
    $\mathcal{D}$                       &The training dataset\\
    $\left( \mathbf{x}, y \right)$      &One instance in $\mathcal{D}$\\
    $M$                                 &The original trained model\\
    $\mathcal{A}(\cdot)$                &The machine learning process\\
    $\mathcal{D}_{u}$                   &The unlearning dataset for unlearning process\\
    $\mathcal{D}_{r}$                   &The remaining dataset after unlearning process\\
    $\mathcal{U}(\cdot)$                &The machine unlearning process\\
    $\mathcal{U}_{t}(\cdot)$            &The target unlearning process\\
    $M_{u}$                             &The model after machine unlearning process\\
    $A_{l}$                             &The input of the $l$-th layer\\
    $f^{k}_{l}$                         &The feature map in $k$-th channel in layer $l$\\
    $G$                                 &The essential graph\\
    $\mathcal{O}(\cdot)$                &The function that used to calculate node value\\
    \hline
	\end{tabular}
\end{table}

\section{Problem Definition}
\label{sec:problemdefinition}

Assume the instance space can be denoted as $\mathcal{X} \subseteq \mathbb{R}^{d}$, with the label space defined as $\mathcal{Y} \subseteq \mathbb{R}$. We use $\mathcal{D} =\left\{\left(\mathbf{x}_{1}, y_{1}\right), \left(\mathbf{x}_{2}, y_{2}\right),...,\left(\mathbf{x}_{n}, y_{n}\right) \right\} \subseteq \mathbb{R}^{d} \times \mathbb{R}$ to represent a training dataset, in which each instance $\mathbf{x} \in \mathcal{X}$ is a $d$-dimensional vector, $y \in \mathcal{Y}$ is the corresponding label, and $n$ is the size of $\mathcal{D}$. Based on this, we can obtain a model $M$ through a machine learning algorithm $\mathcal{A}$, that is, $M = \mathcal{A}(\mathcal{D})$. Other symbols that appear in this paper and their corresponding descriptions are listed in TABLE~\ref{tab:notations}.

\begin{figure*}[!t]
    \centering
    \includegraphics[width=0.9\textwidth]{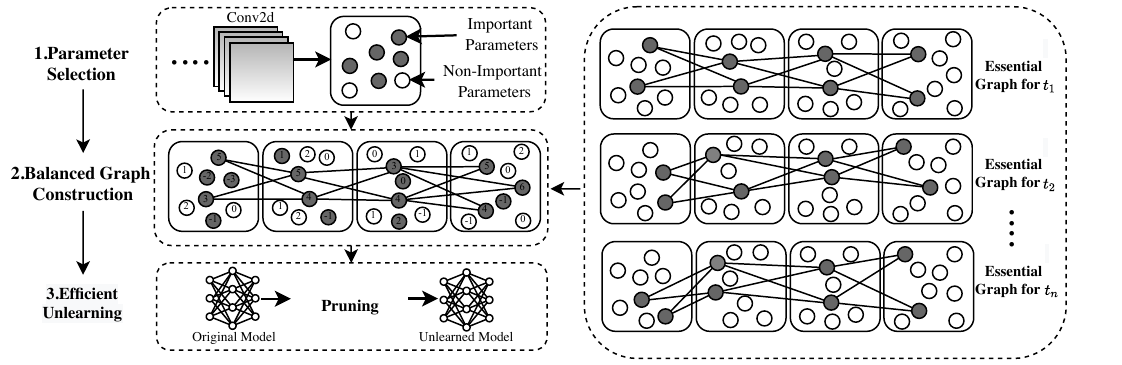}
    \caption{A schematic view of target unlearning.}
    \label{fig:schemeframework}
\end{figure*}

Let $\mathcal{D}_{u} \subset \mathcal{D}$ be a subset of the training dataset, whose influence we want to remove from the trained model $M$. Let its complement $\mathcal{D}_{r}=\mathcal{D}_{u}^{\complement} = \mathcal{D}/\mathcal{D}_{u}$ be the dataset that we want to retain. Now we give the definition of machine unlearning.

\begin{definition}[Machine Unlearning~\cite{DBLP:conf/sp/CaoY15}]
    \label{Definition:Machineunlearning}
    Consider a cluster of instances that we want to remove from the training dataset and the already-trained model, denoted as $\mathcal{D}_{u}$. An unlearning process $\mathcal{U}(M, \mathcal{D}, \mathcal{D}_{u})$ is defined as a function from an already-trained model $M = \mathcal{A}(\mathcal{D})$, a training dataset $\mathcal{D}$, and an unlearning dataset $\mathcal{D}_{u}$ to a model $M_u$, which ensures that the unlearned model $M_{u}$ performs as though it had never seen the unlearning dataset $\mathcal{D}_{u}$.
\end{definition}

Now, we give the definition of target unlearning. which focuses on the unlearning request on the target level: 

\begin{definition}[Target Unlearning]
    Consider one object within instances, denoted as $t$, that we want to remove its effects from all training datasets and the trained model. Target unlearning process $\mathcal{U}_{t}(M, \mathcal{D}, t)$ is defined as a function from an already-trained model $M = \mathcal{A}(\mathcal{D})$, a training dataset $\mathcal{D}$, and a specified object $t$ to a model $M_u$, which ensures that the unlearned model $M_{u}$ performs as though it had never seen the object $t$ in all training dataset.
\end{definition}

\section{Methodology}
\label{sec:methodology}

\subsection{Overview}

As shown in Figure~\ref{fig:schemeframework},  target unlearning mainly consists of the following three steps: parameters selection, balanced graph construction and efficient unlearning, where the balanced graph construction is based on the essential graph data structure for each target. For the purpose of simplification in explanation, we first introduce the concept of an essential graph~(Section~\ref{subsec:explanationgraphandcriticalparameters}). Then, we illustrate how to select the important parameters for a special target within each layer based on the model explanation~(Section~\ref{subsec:parametersselection}). After those steps, we construct the balanced graph to identify parameters that are simultaneously important for the remaining targets and determine which parts of the parameters have a critical influence on unlearning target. After those steps, we manipulate critical parameters that have the greatest effect on the unlearned target to achieve the unlearning purpose~(Section~\ref{subsec:efficientunlearningprocess}).

\subsection{Essential Graph}
\label{subsec:explanationgraphandcriticalparameters}
The core of the model explanation is that model parameters contribute differently to the overall model performance~\cite{DBLP:conf/cvpr/ZhouKLOT16}. This means that some parameters are universally important for all targets, while others are selectively important for specific targets. Appropriate pruning of these parameters can affect the performance of the model with respect to the corresponding targets, which will achieve the purpose of unlearning.

\begin{figure}
    \centering
    \subfloat[The Output of Last layer]{\includegraphics[width=0.25\textwidth]{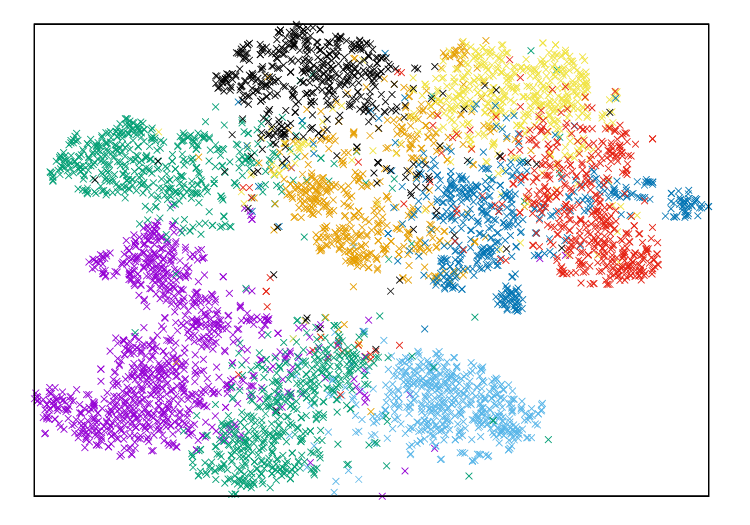}
    \label{fig:theoutputoflastlayer}}
    \subfloat[The Output of Penultimate Layer]{\includegraphics[width=0.25\textwidth]{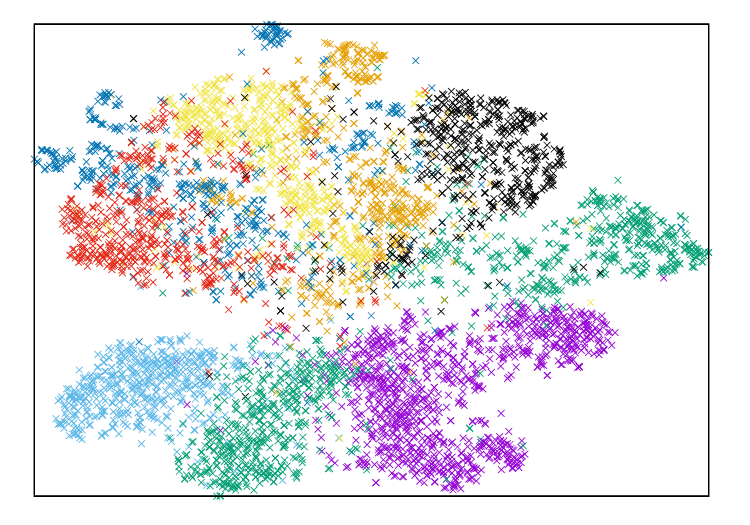}
    \label{fig:theoutputofinterlayer}}
    \caption{Evaluation of data leakage from the penultimate-layer based on STL10 dataset and VGG11 model. }
    \label{fig:evaluationofdataleakage}
\end{figure}

However, it is insufficient to unlearn targets by only considering one-layer parameters~\cite{DBLP:journals/ml/BaumhauerSZ22}. Only selecting the most influential parameters within one layer may not meet the goal of effectiveness due to the specific structure of the deep neural network. For example, if only considering the last layer to select parameters that contain information about targets and then pruning that information from the last layer, once the attacker obtains all parameters of the model in a white-box scenario, they can still deduce information about the unlearning target. As shown in Figure~\ref{fig:evaluationofdataleakage}, we construct an easy-to-understand experiment to show this data leakage problem. We cluster the outputs of the last layer and the penultimate layer of the VGG11 model based on the tSNE algorithm, respectively. From the results, it can be seen that the clustering results are very similar, which indicates that the output of the penultimate layer still contains some information about all data; removing information only at the last layer is not enough. In addition, model parameters in the neighboring layer are usually interactive, considering more than one layer simultaneously can efficiently analyze the impact of parameters on the targets that need to be unlearned.


We assume that the model needs to execute an unlearning process containing $L$ convolutional layers. Each layer's parameters can be represented as $\mathbf{w}_l = \mathcal{R}^{O_{l} \times I_{l} \times K_{l} \times K_{l}}$, where $O_{l}, I_{l}$ and $K_{l}$ denote the number of output channels, number of input channels, and the kernel size, respectively. We use $A_{l} \in$ $\mathcal{R}^{I_{l} \times H_{l} \times W_{l}}$ to denote the input of this convolutional layer, where  $H_{l}$, $W_{l}$ are the height and width of the input respectively. The output feature maps of this layer is calculated as $F_{l}=A_{l} \otimes \mathbf{w}_l$, where $\otimes$ is the convolutional operation, $F_{l} \in \mathcal{R}^{O_{l} \times H_{l+1} \times W_{l+1}}$. We use $f_l^{k}$ to denote the feature maps for the $k$-th channel in this convolutional layer. The output of this CNN model can be represented as $Y = M(\mathbf{x})$. We denote $Y_t$ as the output probability of target $t$. 

To describe and analyze the relationship of important parameters between multiple layers, we construct one useful graph data structure, named \textit{essential graph}. 

\begin{figure}[!t]
    \centering
    \includegraphics[width=0.45\textwidth]{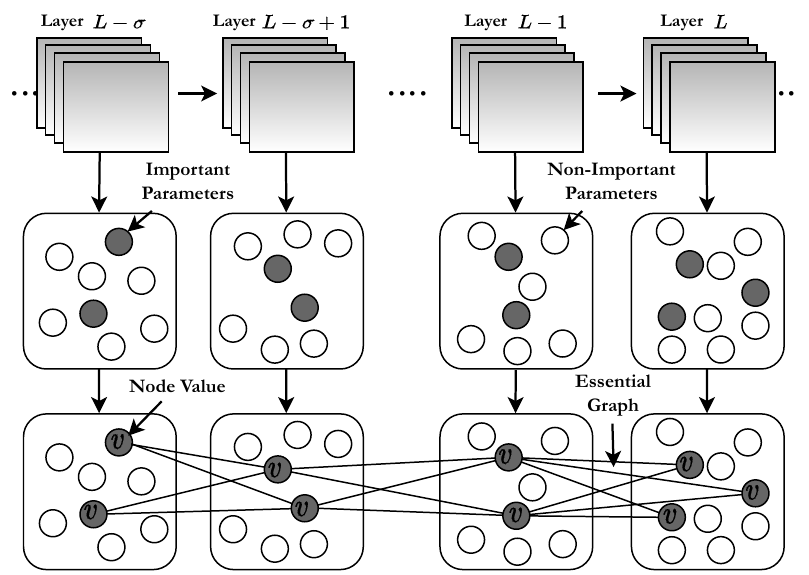}
    \caption{The construction of the essential graph.}
    \label{fig:calculationofeachlayer}
\end{figure}

\begin{definition}[Essential graph]
    An essential graph, $G = (\mathcal{V}, \mathcal{E})$, describes the relationship between important parameters. Each node in $\mathcal{V}$ indicates the importance score of each corresponding output channel, while the edges in $\mathcal{E}$ between each node indicate the connection relationships.
\end{definition}

Specifically, assuming that the set of important channels in layer $l$ for target $t$ can be expressed as $\mathcal{T}_{l}^{t}$~(The calculation of $\mathcal{T}_{l}^{t}$ will be explained later). The essential graph can be denoted as follows:

\begin{equation}
	\begin{split}
	G & = (\mathcal{V}, \mathcal{E})\\
	&s.t.\mathcal{V} = \left\{v_{l,k}|~k \in  [0,|\mathcal{T}_{l}^{t}|], v_{l,k} = \mathcal{O}(\mathcal{T}_{l,k}^{t}) \right\}\\
        &~~~~\mathcal{E} = \left\{e_{(l,i),(l+1,j)}|~i \in  [0,|\mathcal{T}_{l}^{t}|], j \in  [0,|\mathcal{T}_{l+1}^{t}|]\right\}\\
        &~~~~l \in [L - \sigma, L - 1]
	\end{split}
\end{equation}

Each element in $\mathcal{T}_{l}^{t}$ is denoted as one channel $\mathcal{T}_{l,k}^{t}$, where $k$ ranges from $0$ to $|\mathcal{T}_{l}^{t}|$, the size of the set $\mathcal{T}_{l}^{t}$. In subsequent sections, when we refer to parameters, it typically denotes all the parameters within one channel. In the essential graph, $v_{l,k}$ represents the value of each node, which is determined by a function $\mathcal{O}(\cdot)$ using $\mathcal{T}_{l,k}^{t}$ as the parameter. The function $\mathcal{O}(\cdot)$ can be arbitrary, but it must ensure that the node values for channels of different importance should be different and values for channels of similar importance are similar. The exact definition of $ \mathcal{O}(\cdot)$ will be provided later. $e_{(l,i),(l+1,j)}$ mean one connection relationship between node $v_{l,i}$ and $v_{l+1,j}$, where $i \in  [0,|\mathcal{T}_{l}^{t}|]$ and $j \in  [0,|\mathcal{T}_{l+1}^{t}|]$.

$\sigma$ is used to control how many layers should be considered. The reason we only consider the last $\sigma$ layers is that the CNN model usually consists of a low-level feature extractor and a high-level classifier. The high-level classifier is used to distinguish different targets, and only removing the information contained in the high-level classifier will generally not influence model performance for the remaining targets.

The edge between two nodes with larger node values indirectly indicates a strong correlation between the two nodes' parameters. For example, the parameters of a layer $l$ for the detection of textures and materials may be jointly linked to a parameter of a particular scene detector in layer $l+1$. All the links between the two layers with larger node values form a route for the transmission of important features of one target.

Algorithm~\ref{algorithm:explanationgraph} performs the construction of \textit{essential graph} for one instance with target label $t$, while Figure~\ref{fig:calculationofeachlayer} shows the process. 
Before this algorithm, the model provider indexes all layers that need to be considered to remove information about a target $t$ and select important parameters w.r.t target $t$. Assume that the results of layer $l$ denoted as $\mathcal{T}_{l}^{t}$, and that in layer $l + 1$ denoted as $\mathcal{T}_{l+1}^{t}$. We use $v_{l,k}$ to denote one node in $G$ of corresponding channel $k$ in layer $l$ and use $\mathcal{O}(\mathcal{T}_{l,k}^{t})$ to represent the node value. The link between use $v_{l,i}$ and $v_{l+1,j}$ is denoted as $e_{(l,i),(l+1,j)}$. For each layer, model provider sequentially add the node $v_{l+1,j}$ and $v_{l,i}$ to $\mathcal{V}$, and add the edge $e_{(l,i),(l+1,j)}$ to $\mathcal{E}$ to construct the relationship between layer $l$ and $l+1$~(Line 2-7). It is important to note that the above steps can be performed at any time during the inference phase; thus, the time consumption of the unlearning process is not increased.

\begin{algorithm}[!t]
	\small
	\caption{Essential Graph Construction}
 	\label{algorithm:explanationgraph}
	\LinesNumbered 
	\KwIn{model $M$, parameter index $\mathcal{T}_{l}^{t}$ with $l$ $\in$ $[L- \sigma, L]$, layer selection factor $\sigma$}
	\KwOut{essential graph:$G$}
  	~~~~~Initialize $\mathcal{V}$ and $\mathcal{E}$ to empty $\emptyset$\\
        ~~~~~\For{\rm each layer $l$ in $[L- \sigma, L - 1]$}{
        ~~~~~\For{\rm each index $\mathcal{T}_{l,i}^{t}$ in $\mathcal{T}_{l}^{t}$}{
        ~~~~~\For{\rm each index $\mathcal{T}_{l+1,j}^{t}$ in $\mathcal{T}_{l+1}^{t}$}{
            $\mathcal{V}$ $\leftarrow$ add $v_{l+1,j}$ with value $\mathcal{O}(\mathcal{T}_{l+1,j}^{t})$  \\
            $\mathcal{E}$ $\leftarrow$ add $e_{(l,i),(l+1,j)}$ \\
            
        }
        $\mathcal{V}$ $\leftarrow$ add $v_{l,i}$ with value $\mathcal{O}(\mathcal{T}_{l,i}^{t})$  \\
        }
        }
	\Return {$G = (\mathcal{V}, \mathcal{E})$}\\
\end{algorithm}

\subsection{Parameters Selection}
\label{subsec:parametersselection}

In the above section, we describe how to construct an essential graph based on the important parameters of each layer. In this section, we describe how to find those important parameters within each layer based on model explanation. 


There are many works to select important parameters within a model, such as ablation methods~\cite{DBLP:journals/corr/abs-1806-02891,DBLP:journals/corr/abs-1901-08644,DBLP:journals/tnn/AmjadLG22}. Ablation methods usually investigate the effect of one unit by removing this unit to understand the contribution of this unit to the overall performance. However, there are two main issues when using these methods to find important parameters for one specific target. Firstly, these methods require indexing all parameters, and the computational cost for complex models with large datasets is high. This will reduce the efficiency of the unlearning process. In addition, these methods analyze the influence of a single parameter at the instance level, which is not sufficient for finding important parameters at the target level. CAM-based methods are another interpretive methods and usually used to highlight the important regions in one image~\cite{DBLP:conf/cvpr/ZhouKLOT16,DBLP:conf/iccv/SelvarajuCDVPB17,DBLP:conf/wacv/ChattopadhyaySH18,DBLP:conf/cvpr/WangWDYZDMH20,DBLP:journals/tip/JiangZHCW21}. It did not originally apply to finding importance parameters. Here, we illustrate how it can be used to analyze the importance for a particular target.

Consider a convolutional layer $l$ in a model $M$. CAM $L_{\text {CAM}}^t$ with respect to target $t$ can be described as follows:

\begin{equation}
    \label{equation:1}
   L_{\text {CAM}}^t=\operatorname{ReLU}\left(\sum_k \alpha_{l}^k f_l^{k}\right)
\end{equation}
where $k$ denotes the index of the feature map in layer $l$. $\alpha_{l}^k f_l^{k}$ denotes the linear combination between $\alpha_{l}^k$ and $f_l^{k}$. $\alpha_{l}^k$ represents the importance of each feature map $f_l^{k}$ for instance with target $t$ and calculated based on various methods, such as Original CAM~\cite{DBLP:conf/cvpr/ZhouKLOT16}, Score-CAM~\cite{DBLP:conf/cvpr/WangWDYZDMH20} and Grad-CAM~\cite{DBLP:conf/iccv/SelvarajuCDVPB17}. For convenience, we calculate $\alpha_{l}^k$ the based on the method in Grad-CAM~\cite{DBLP:conf/iccv/SelvarajuCDVPB17}:
\begin{equation}
    \label{equation:2}
    \alpha_{l}^k=\frac{1}{Z} \sum_i \sum_j \frac{\partial Y_t}{\partial f_{l,(i,j)}^k}
\end{equation}
where $\frac{1}{Z} \sum_i \sum_j$ in Equation~\ref{equation:2} represents the global average pooling operation, while $\frac{\partial Y_t}{\partial f_{l,(i,j)}^k}$ denotes the gradients of the $Y_t$ with respect to $f_{l,(i,j)}^k$. 

Based on the above steps, Grad-CAM will generate a heatmap that can highlight the important regions for a given target in one image. Next, we will simply illustrate that $\alpha_{l}^k$ also represents the importance of each channel for all instances with the same target. We first give the definition of the influence of channel parameters:

\begin{algorithm}[!t]
	\small
	\caption{Importance Parameter Selection}
 	\label{algorithm:importantparametercalculation}
	\LinesNumbered 
	\KwIn{model $M$, target label $t$, layer $l$}
	\KwOut{index of important parameters: $\mathcal{T}_{l}^{t}$}
	~~~~~Initialize $\mathcal{Z}_{l}$ to an empty directory $\emptyset$\\
 	~~~~~Initialize $\mathcal{T}_{l}^{t}$ to an empty directory $\emptyset$\\
	~~~~~Select and feed a instance $\mathbf{x}$ to model $M(\mathbf{x})$\\
        ~~~~~\For{\rm each channel $k$ in layer $l$}{
            ~~~~~Calculate $\alpha_{l}^k=\frac{1}{Z} \sum_i \sum_j \frac{\partial Y_t}{\partial f_{l,(i,j)}^k}$\\
            ~~~~~Add $\alpha_{l}^k$ to $\mathcal{Z}_{l}$\\
	}
        ~~~~~\For{\rm each channel $k$ in layer $l$}{
            ~~~~~\If{$\alpha_{l}^k$ in Top $\delta$}{
                ~~~~~$\mathcal{T}_{l,k}^{t}$ = True \\
                }
            ~~~~~\Else{
                ~~~~~$\mathcal{T}_{l,k}^{t}$ = False \\
                }
	}
 	\Return {$\mathcal{T}_{l}^{t}$}\\
\end{algorithm}

\begin{definition}(\textit{Influence of Channel Parameters}).
    \label{def:effectofindividual}
    Given a model $M$ that takes an instance $\mathbf{x}$ with target $t$. It will generate the heatmap with the intermediate weights denoted as $\alpha_{l}^k$. We define the influence of corresponding channel parameters $w_{l}^{k}$ in layer $l$ toward target $t$ as $s_{w_{l}^{k}} = \alpha_{l}^k$.
\end{definition}

Consider a CNN model with $L$ layers, excluding the final linear layers; the model output can be expressed as $Y = R(\mathbf{w}_l \otimes R(\mathbf{w}_{l-1} \otimes \ldots \otimes R(\mathbf{w}_2 \otimes R(\mathbf{w}_1 \otimes x)) \ldots ))$, where $\mathbf{w}_{l}$ represents the parameter in layer $l$,  $R(\cdot)$ is the activation function, and $\mathbf{x}$ is the model input. We already know that each $\alpha_{l}^k$ represents the importance of the corresponding feature map $f_l^{k}$. We use the function $I(\cdot)$ to denote importance, and denote the relationship mentioned above as $\alpha_{l}^k \propto I(f_l^{k})$. 

Consider one layer $l$ within the model, with input denoted as $A_{l} \in \mathcal{R}^{I_{l} \times H_{l} \times W_{l}}$, where $A_{l} = R(\mathbf{w}_{l-1} \otimes \ldots \otimes R(\mathbf{w}_2 \otimes R(\mathbf{w}_1 \otimes x)) \ldots )$, and parameters as $\mathbf{w}_l$. The output of this layer, that is, feature maps, can be denoted as $F_l = A_{l} \otimes \mathbf{w}_l$, where $f_l^{k} = A_{l}^{k} \otimes \mathbf{w}_l^{k}$. When changing one specific channel's parameters $w_{l}^{k}$ in layer $l$ to obtain $w_{l}^{'}$, the output will be changed to $F_{l}^{'} = A_{l} \otimes w_{l}^{'}, F_{l} \neq F_{l}^{'}$. This means that when given an instance, the output feature maps of this layer are only determined by the parameters in this layer. We use the notation $\rightarrow$ to denote this relationship and define it as $w_{l}^{k} \rightarrow  f_l^{k}$. 

Based on the above two inferences, namely $\alpha_{l}^k \propto I(f_l^{k})$ and $w_{l}^{k} \rightarrow  f_l^{k}$, we can deduce $\alpha_{l}^k \propto I(w_{l}^{k}) = s_{w_{l}^{k}}$ in definition~\ref{def:effectofindividual}. This means $\alpha_{l}^k$ represents the importance of the output feature map~\cite{DBLP:conf/cvpr/WangWDYZDMH20,DBLP:conf/cvpr/ZhouKLOT16,DBLP:conf/iccv/SelvarajuCDVPB17,DBLP:journals/tkde/LiCSBGQWGZXC22}, it also represents the importance of the corresponding channel parameters of this layer. In addition, based on the model explanation work in~\cite{DBLP:journals/corr/abs-1806-02891,DBLP:journals/corr/abs-1901-08644,DBLP:journals/tnn/AmjadLG22}, we can conclude that given two different instances with the same target label $t$, $\mathbf{x}_i^{t}$ and $\mathbf{x}_j^{t}$, the distribution $\mathbb{E}(\cdot)$ of influence of channel parameters in layer $l$ for two instances, $\mathbb{E}(s_{w_{l}^{k}}, \mathbf{x}_i^{t})$ and $\mathbb{E}(s_{w_{l}^{k}}, \mathbf{x}_j^{t})$, is similar. Based on this, we can select the most important channel parameters for the unlearning target within one layer based on one instance, then use reasonable operations to remove the information in those parameters.

The process of importance parameter selection is described in  Algorithm~\ref{algorithm:importantparametercalculation}. In line 2, in order to calculate the importance within a layer for a target $t$, the model provider first selects an instance $\mathbf{x}$ that contains one or a group of targets, one of which's target label is $t$. Then, this selected instance will be fed to the model $M$ to record all essential data to calculate each channel's $\alpha_{l}^k$~(lines 3-6). After that, lines 7-11 return the index of selected important parameters, where $\delta$ is a hyper-parameter used to control the proportion of important parameters.

\subsection{Balanced Graph Construction and Unlearning}
\label{subsec:efficientunlearningprocess}

Given the essential graph generated based on Section~\ref{subsec:explanationgraphandcriticalparameters} using the important parameters within each layer selected based on Section~\ref{subsec:parametersselection}, it is worth noting that some task-specific parameters in this graph, such as those used to detect textures, are not only important for the unlearning target but may also be used for other detection tasks~\cite{DBLP:conf/www/Wang0XQ22}. For example, the parameters for target $t$ used to detect textures or materials may also be used to detect other targets' textures or materials. Directly removing information on all those parameters will destroy the performance of other targets in the remaining data. To balance the unlearning effectiveness and model performance of remaining targets. We further select the most critical parameters based on balance operations. 

Our main idea is to simultaneously consider both unlearning target and remaining targets in the process of constructing graph nodes, and set different node values according to the type of target. After that, summing these node values, and use the final summation to reflect the effect of the parameters on unlearning target. For the calculation of the values $\mathcal{O}(\mathcal{T}_{l,k}^{t})$, we define as:

\begin{equation}
        \label{equ:nodevalue}
	\mathcal{O}(\mathcal{T}_{l,k}^{t}) =\\
        \begin{cases}
	|Y|,   & \mathcal{T}_{l,k}^{t} = True, t \in \mathcal{D}_u\\
	-1,    & \mathcal{T}_{l,k}^{t} = True, t \in \mathcal{D}_r\\
        0,     & \mathcal{T}_{l,k}^{t} = False\\
       \end{cases}
\end{equation}

where $|Y|$ denotes the number of targets in training data. The above equation indicates that we consider different ways of calculating node values in constructing the essential graph; for unlearning target, we calculate based on $\mathcal{O}(\mathcal{T}_{l,k}^{t}) =|Y|$  for each selected channel, while for targets from the remaining data, we calculate based on $\mathcal{O}(\mathcal{T}_{l,k}^{t}) = -1$. For all channels that are not selected as important channels, we calculate based on $\mathcal{O}(\mathcal{T}_{l,k}^{t}) = 0$. After each target has constructed its corresponding essential graph, we sum the graphs constructed by different targets. If a graph node is important for both data, then the summed node value $v_{l,k}^{sum} = |Y|_{t \in \mathcal{D}_{u}} + \sum_{t \in \mathcal{D}_{r}}({-1})$ will tend to the target number of $D_u$. If the node is important for the unlearning target only, the node value will tend to $v_{l,k}^{sum} =|Y|_{t \in \mathcal{D}_{u}}$.

Algorithm~\ref{algorithm:explanationgraphconstruction} describes the balancing process of the essential graph. In lines 1-4, the model provider first select important parameters for each target within each layer in $[L- \sigma, L]$. $\mathcal{T}^{t}_{l}$ represents the result of target $t$ in layer $l$ and $\mathcal{T}^{t}$ represents the result of target $t$ in all layers. Then, the model provider constructs each essential graph based on the Algorithm~\ref{algorithm:explanationgraph}, in which the model provider will choose different $\mathcal{O}(\mathcal{T}_{l,k}^{t})$ in Equation~\ref{equ:nodevalue} based on the type of target~(Lines 5-6). Lines 7-9 accumulate the values of all the corresponding nodes to get the balanced essential graph.

\begin{algorithm}[!t]
	\small
	\caption{Balanced Graph Construction}
 	\label{algorithm:explanationgraphconstruction}
	\LinesNumbered 
	\KwIn{model $M$, training data $\mathcal{D}$}
	\KwOut{balanced essential graph:~$G^{*}$}
        ~~~~~\For{\rm each target $t$ in $\mathcal{D}$}{
        ~~~~~\For{\rm each layer $l$ in $[L- \sigma, L]$}{
              ~~~~~$\mathcal{T}_{l}^{t}$ $\leftarrow$ calculate based on Algorithm~\ref{algorithm:importantparametercalculation}~($M$, $t$, $l$)\\
              ~~~~~Add $\mathcal{T}_{l}^{t}$ to $\mathcal{T}^{t}$ \\
              }
	}
        ~~~~~\For{\rm each target $t$ in $\mathcal{D}$}{
              ~~~~~$G^{t}$ $\leftarrow$ construct based on Algorithm~\ref{algorithm:explanationgraph}($M$, $\mathcal{T}^{t}$)\\

        }
        ~~~~~\For{\rm each target $t$ in $\mathcal{D}$}{
              ~~~~~$v_{l,k}^{sum}$ $\leftarrow$ Accumulate all node value with the same layer $l$ and channel $k$ in each $G^{t}$\\
              ~~~~~Add $v_{l,k}^{sum}$ with edges to $G^{*}$
        }
	\Return {$G^{*}$}\\
\end{algorithm}

After the balance operation, the operation of target unlearning is simplified to selecting appropriate nodes based on the nodes' values in $G^{*}$ and manipulating the model parameters corresponding to these selected nodes to remove information about unlearning targets. As shown in Algorithm~\ref{algorithm:unlearningprocess},  for each selected layer, we perform the pruning operation for channels where the corresponding node value is equal to $|Y|$. 

\begin{algorithm}[!t]
	\small
 	\caption{Unlearning Process}
	\label{algorithm:unlearningprocess}
	\LinesNumbered 
	\KwIn{Balanced Graph $G^{*}$, Model $M$}
	\KwOut{unlearned model $M_{u}$}
        ~~~~~$M_{u} \leftarrow M$\\
        ~~~~~\For{\rm each layer $l$ in $[L- \sigma, L]$ within $M_{u}$}{
        ~~~~~\For{\rm each $v_{l,k}$ in $G^{*}$}{
            ~~~~~\If{$v_{l,k}$ is equal to $|Y|$}{
                  ~~~~~Pruning $w_{l}^{k}$ from the model.\\
                }
            }
        }
	\Return {$M_{u}$}\\
\end{algorithm}

\subsection{Performance Analysis}

In this section, we analyze the performance of the proposed scheme in terms of computational, and storage overheads. 

\subsubsection{Computational Overhead}

The model training process is the most time-consuming element of machine learning. In the interests of simplicity, we set $f(\cdot)$ to represent the computational cost of the forward propagation, then the computational cost of one-step backpropagation is at most $5f(\cdot)$~\cite{DBLP:conf/infocom/LiuXYWL22}.  

Our computations primarily revolve around identifying important parameters, specifically in the Grad-CAM calculation process. This process will involve a single forward propagation and $\sigma$ backpropagation for one target. Therefore, when selecting the important parameters of a target in the selected layer, it requires $6f(\cdot)$.  Moreover, consider that we need to construct the balanced essential graph, and this process involves all unlearning targets and the remaining targets within the training dataset. Therefore, neglecting basic arithmetic operations, the complexity of our approach is approximately $6f(\cdot) * |Y|$.  This value is much smaller than the cost of model training, which will involve multiple epochs and batch sizes, and it's about $\mathcal{N}_{epoch} * \mathcal{N}_{batch} * \text{batch size} * 6f(\cdot)$.

\subsubsection{Storage Overhead}

The storage consumed by our target unlearning scheme primarily involves storing the results of Grad-CAM. We use $\mathbf{w}_l = \mathcal{R}^{O_{l} \times I_{l} \times K_{l} \times K_{l}}$ to denote one layer's parameters, where $O_{l}, I_{l}$ and $K_{l}$ denote the number of output channels, number of input channels, and the kernel size, respectively. Since we consider the importance of different channels in each layer, storing this importance will cost $O_{l}$ for one target. Due to the need to consider all targets within the dataset and select multiple layers in the construction process of the balanced essential graph, our storage cost is approximately $\sigma * |Y| * O_{l}$, excluding basic arithmetic operations. Compared to the storage cost of model parameters, which is typically $L * O_{l} \times I_{l} \times K_{l} \times K_{l}$, our storage requirement is very small.

\textit{Summary}: From the above analysis, the proposed unlearning schemes are determined to be efficient in terms of computational and storage costs, which demonstrates the practical potential and significant performance improvements obtained by our unlearning schemes.

\begin{table*}[]
\caption{Experimental settings in evaluating the performance of multi-classification.}
\label{tab:experimentsetting}
\centering
    \renewcommand{\arraystretch}{1.3}
    \begin{tabular}{ccccc}
    \hline
    Different settings &  Original targets        & Target that need to be unlearned    & $\delta$  & $\sigma$ \\ \hline
    I         &Bald, Mouth Slightly Open                  & Bald                                & 0.1       & 5   \\
    II        &Bald, Mouth Slightly Open                  & Mouth Slightly Open                 & 0.1       & 5     \\
    III       &Mouth Slightly Open, No Beard, Eyeglasses  & Eyeglasses                          & 0.08      & 5     \\
    IV        &Smiling, No Beard, Eyeglasses              & No Beard,Eyeglasses                 & 0.1       & 5    \\\hline
    \end{tabular}
\end{table*}

\begin{table*}[]
\caption{Target unlearning results for multi-classification~(\%).}
\label{tab:experimentresultsforperformance}
\centering
    \renewcommand{\arraystretch}{1.3}
    \begin{tabular}{cccccccc}
    \hline
                                 &                     & Original & Retraining & Chundawat et al.~\cite{DBLP:conf/aaai/ChundawatTMK23}     & Ayush et al.~\cite{TNNLSAyush2023}     & Foster et al.~\cite{DBLP:conf/aaai/FosterSB24}    & Our   \\ \hline
        \multirow{2}{*}{I}      & Bald                & 67.55    & 0.00                 & 0.00 & 0.00 & 0.00 & 0.00 \\
                                & Mouth Slightly Open & 92.50    & 0.00                 & 0.00 & 0.00 & 0.00 & 91.56 \\
        \multirow{2}{*}{II}     & Bald                & 67.55    & 0.00                 & 0.00 & 0.00 & 0.00 & 66.56 \\
                                & Mouth Slightly Open & 92.50    & 0.00                 & 0.00 & 0.00 & 0.00 & 0.00 \\
        \multirow{3}{*}{III}    & Mouth Slightly Open & 90.19    & 0.00                 & 0.00 & 0.00 & 0.00 & 90.07 \\
                                & No Beard            & 96.11    & 0.00                 & 0.00 & 0.00 & 0.00 & 95.78 \\
                                & Eyeglasses          & 97.18    & 0.00                 & 0.00 & 0.00 & 0.00 & 0.00  \\
        \multirow{3}{*}{IV}     & Smiling             & 85.56    & 0.00                 & 0.00 & 0.00 & 0.00 & 83.06 \\
                                & No Beard            & 96.73    & 0.00                 & 0.00 & 0.00 & 0.00 & 0.00 \\
                                & Eyeglasses          & 97.43    & 0.00                 & 0.00 & 0.00 & 0.00 & 0.00  \\ \hline
    \end{tabular}
\end{table*}

\begin{table}[]
\caption{Computational cost associated with target unlearning for multi-classification~(s).}
\label{tab:experimentresulstime}
\centering
    \renewcommand{\arraystretch}{1.3}
    \begin{tabular}{ccccc}
    \hline
    Different settings   & Graph Generation Cost               & Unlearning Cost   \\ \hline
    I          & 34.24                               & 0.03         \\
    II         & 33.21                               & 0.03           \\
    III        & 123.13                             & 0.03         \\
    IV         & 71.35                              & 0.05          \\\hline
    \end{tabular}
\end{table}

\section{Performance Evaluation}
\label{sec:experiment}

\subsection{Experiment Setup}
\subsubsection{Model and Dataset}
We choose three popular models, including AlexNet, VGG and ResNet, and use five widely used public image datasets: MNIST~\footnote{http://yann.lecun.com/exdb/mnist/}, CIFAR-10 and CIFAR-100~\footnote{https://www.cs.toronto.edu/~kriz/cifar.html} and ImageNet ILSVRC2012~\footnote{https://www.image-net.org/} and CelebA~\footnote{https://mmlab.ie.cuhk.edu.hk/projects/CelebA.html} to evaluate our scheme. The datasets cover different attributes, dimensions, and numbers of categories, allowing us to explore the unlearning utility of the proposed algorithm effectively.

\subsubsection{Baseline Methods} To demonstrate the effectiveness of our scheme, we consider the following baseline schemes:

\begin{itemize}
    \item Retraining from scratch~\cite{DBLP:conf/sp/CaoY15}: The most straightforward way for machine unlearning involves retraining the model from scratch, after deleting the instances that need to be unlearned from the training dataset. Thus, for comparison, we retrain the model from scratch after deleting all instances that contain targets. Normally, the model retrained from scratch is the optimal unlearned model.  
    \item Knowledge Distillation: Chundawat et al.~\cite{DBLP:conf/aaai/ChundawatTMK23} proposed a novel teacher-student framework with knowledge distillation for unlearning, where both competent and incompetent teachers transfer knowledge to a student model. This method leverages the incompetent teacher to unlearn specific data while retaining general knowledge through the competent teacher. 
    \item Impair and Repair: Ayush et al.~\cite{TNNLSAyush2023} presented impair and repair-based unlearning method. They first generated an error-maximizing noise matrix that contained highly influential patterns corresponding to the unlearning class. This matrix is then used to fine-tune model for unlearning. After impairing, further fine-tuning with the remaining data is performed to maintain model performance.
    \item Pruning-Based: Foster et al.~\cite{DBLP:conf/aaai/FosterSB24} proposed a unlearning method called Selective Synaptic Dampening (SSD). First, SSD used the Fisher Information Matrix (FIM) to identify parameters most influenced by the data to be forgotten. Then, SSD induces unlearning by dampening these identified parameters proportionally to minimize their impact, thereby unlearning the specified data while maintaining overall model performance.
\end{itemize}

In Section~\ref{subsec:efficientunlearningprocess}, we construct a balanced graph to minimize the impact on the remaining data when unlearning targets. To illustrate the effectiveness of this strategy, we also consider the following methods for comparison.
\begin{itemize}
    \item Unlearning within last-layer: For this setting, we only select and prune the important parameters within the last layer for unlearning targets.
    \item Unlearning without balance: For this setting, we select and prune the important parameters based on the essential graph without balance operation.
\end{itemize}

\subsubsection{Metrics}
Training models contain various randomness, especially for complex deep models with enormous training datasets. It is difficult to determine if the unlearning scheme has effectively eliminated the impacts of targets. For evaluating our scheme, we consider two aspects, including performance-based and attack-based metrics.

\begin{itemize}
    \item \textbf{Performance-based}: Generally, trained models often exhibit high performance for training data. Therefore, the unlearning process could be verified by the performance of the model. For targets that need to be unlearned, if the performance ideally is the same as that of a model trained without seeing the unlearning targets, it can indicate that the unlearning process satisfies \textit{effectiveness}. If model performance for remaining data is almost kept unchanged after the unlearning process, it means that the unlearning procedure achieves the \textit{model utility} goal.

    \item \textbf{Attack-based}: The basic purpose of unlearning is to reduce the risk of sensitive information leakage. Therefore, certain attack methods can be used to directly and effectively verify the success of unlearning operations. Here, we use model inversion attack~\cite{DBLP:conf/ccs/FredriksonJR15} and membership inference attacks~(MIAs)\cite{DBLP:conf/csfw/YeomGFJ18} to evaluate our scheme. For MIAs, we utilize the recall = $\frac{TP}{Tp + FN}$ to represent the results of our evaluation. 
    
\end{itemize}

\subsection{Performance Analysis}
To evaluate the effectiveness of our unlearning scheme, we do the following experiments:

\begin{figure}[!t]
      \centering
      \subfloat[Before Unlearning~(Person)]{\includegraphics[width=0.25\textwidth]{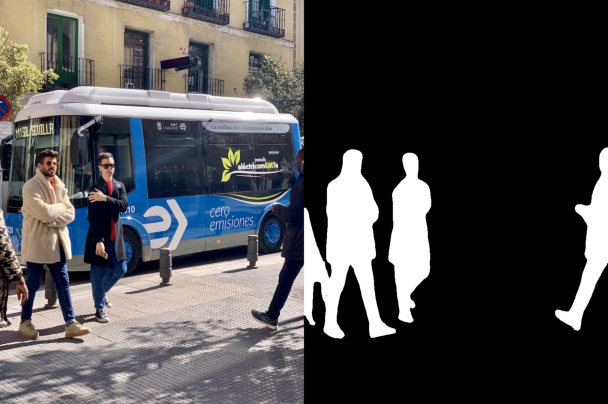}}
      \subfloat[Before Unlearning~(Bus)]{\includegraphics[width=0.25\textwidth]{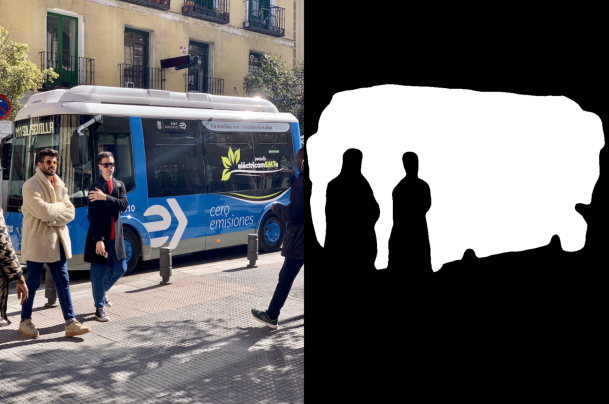}}\\
      \subfloat[After Unlearning~(Person)]{\includegraphics[width=0.25\textwidth]{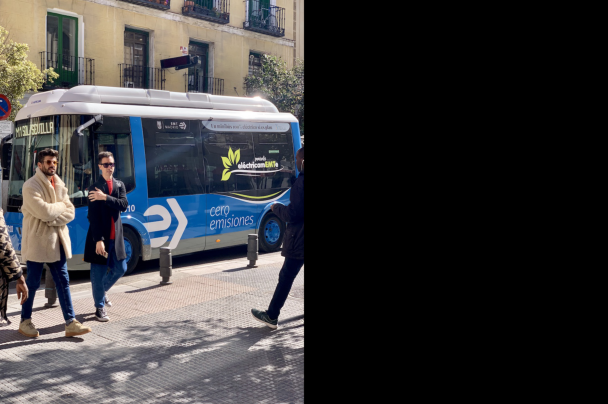}}
      \subfloat[After Unlearning~(Bus)]{\includegraphics[width=0.25\textwidth]{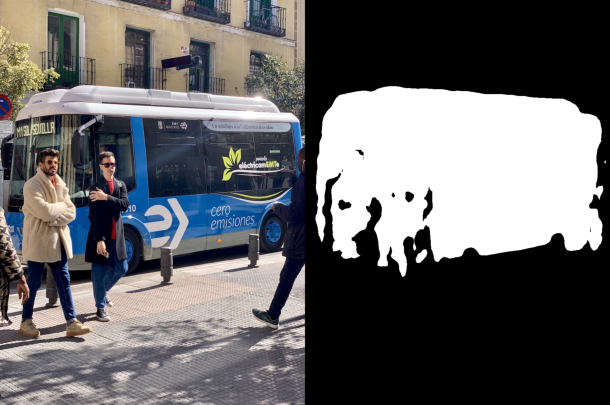}}
    \caption{Target unlearning results for semantic segmentation.}
    \label{fig:semanticsegmentation}
\end{figure}
\begin{figure}[!t]
      \centering 
      \subfloat[Instance~(1) Before Unlearning]{\includegraphics[width=0.25\textwidth]{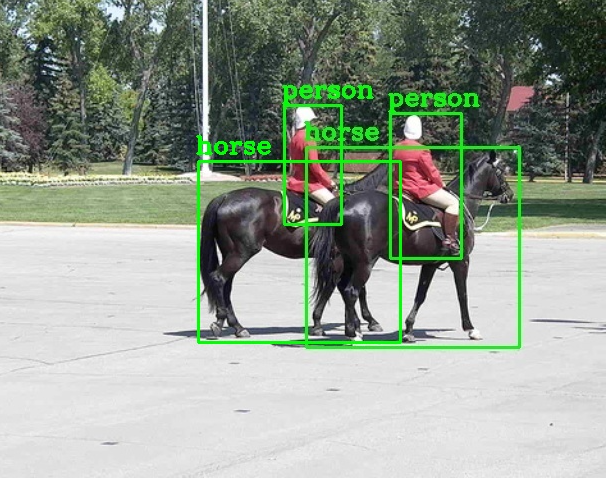}}
      \subfloat[Instance~(1) After Unlearning]{\includegraphics[width=0.25\textwidth]{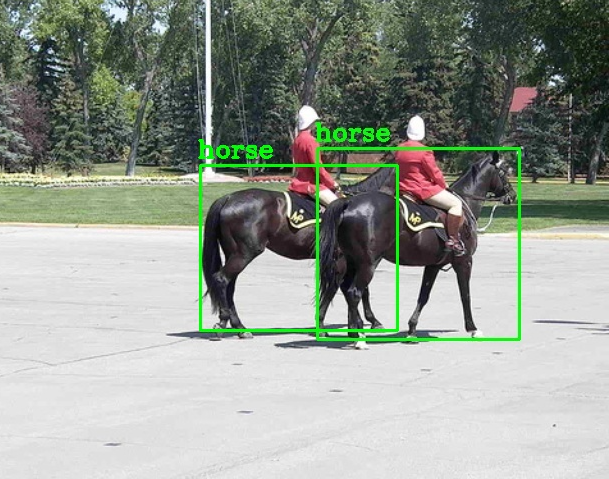}}\\
      \subfloat[Instance~(2) Before Unlearning]{\includegraphics[width=0.25\textwidth]{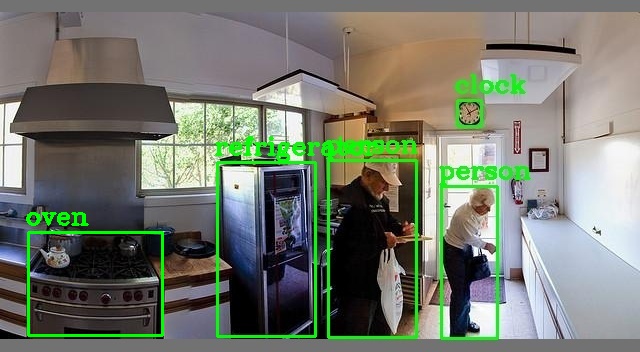}}
      \subfloat[Instance~(2) After Unlearning]{\includegraphics[width=0.25\textwidth]{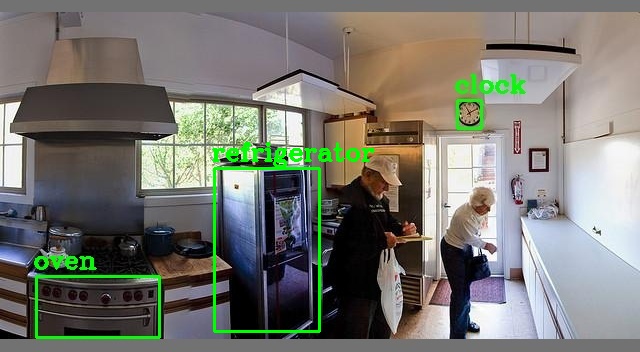}}\\
    \caption{Target unlearning results for object detection.}
    \label{fig:objectdetection}
\end{figure}

\subsubsection{Evaluating Performance of Multi-Classification Task} We first consider a multi-classification task based on CelebA dataset as a scenario containing multiple targets, where each class, such as nose, mouth, and ears, is considered a different target. This configuration enhances the comprehension of the novelty in our paper. Each image in CelebA contains all targets, such as the nose and mouth. When unlearning a target, existing unlearning schemes, due to their coarse granularity, can only remove all instances to achieve unlearning goals, thereby affecting the performance of the model after unlearning. Our target unlearning scheme can effectively remove targets, such as mouth information while maintaining the model's ability to recognize other targets.

We choose the ResNet as our model and consider various unlearning settings as shown in Table~\ref{tab:experimentsetting}, including scenarios involving the unlearning of one and two targets. For all original model training processes, we set epoch = 20, batch size = 128, and learning rate = 5e-06. We also compare our scheme with other four existing schemes, including retraining from scratch, knowledge distillation~(Chundawat et al.~\cite{DBLP:conf/aaai/ChundawatTMK23}), impair and repair~(Ayush et al.~\cite{TNNLSAyush2023}) and pruning-base~(Foster et al.~\cite{DBLP:conf/aaai/FosterSB24}). Experimental results are shown in Table~\ref{tab:experimentresultsforperformance}, while the corresponding computational cost is shown in Table~\ref{tab:experimentresulstime}.

From all the results, our scheme can effectively unlearn all target-related information from the model, making the unlearned model unable to recognize the corresponding targets. For example, as illustrated in setting II, after the unlearning process, the accuracy of if the mouth is slightly open decreased from $92.50\%$ to $0\%$. As shown in setting IV, our approach also remains effective for simultaneously unlearning two or more targets. However, for retraining from scratch, Knowledge Distillation~(Chundawat et al.~\cite{DBLP:conf/aaai/ChundawatTMK23}), Impair and Repair~(Ayush et al.~\cite{TNNLSAyush2023}) and Pruning-Base~(Foster et al.~\cite{DBLP:conf/aaai/FosterSB24}). the unlearning results are quite poor. For example, as shown in setting II, when unlearning \textit{mouth slightly open}, the performance of the remaining data also decreases to $0\%$. This is due to the fact that all the above unlearning schemes lack a more fine-grained unlearning strategy, operating only at the instance level. When all instances contain this unlearning target, information related to other targets within that instance is also removed, leading to a decline in model performance. Take retraining from scratch as an example; when considering the unlearning of one target from the model, such as the information of bald, the only way of retraining from scratch is to remove all the instances containing bald and then retrain the model from scratch. In cases where all instances in the dataset contain the information that needs to be unlearned, such as in the CelebA dataset, this would result in the deletion of all instances, leaving no dataset available for model retraining. Consequently, the final model performance will be $0\%$. 

From Table~\ref{tab:experimentresulstime}, it can be seen that the cost required for our scheme is very short, with all costs being less than 0.1s. Although the construction of the graph consumes more time, it is still much lower compared to retraining from scratch, which takes approximately 4 hours and 30 minutes. Moreover, the graph construction can be performed separately before unlearning, so it will not cost time in the unlearning process.

\begin{figure}[!t]
    \centering
    \subfloat[Original Model]{\includegraphics[width=0.48\textwidth]{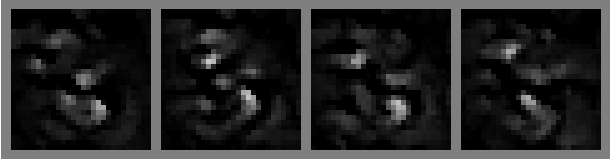}}\\
    \subfloat[Fully Retraining]{\includegraphics[width=0.48\textwidth]{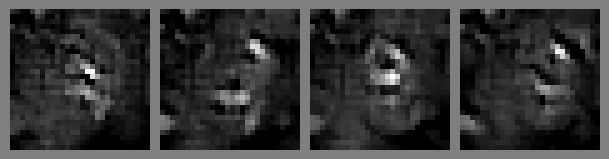}}\\
    \subfloat[Target Unlearning]{\includegraphics[width=0.48\textwidth]{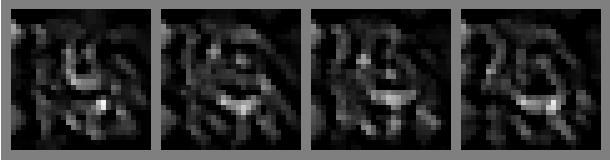}}
    \caption{The results of model inversion attack.}
\label{fig:unlearningdataaccuracy}
\end{figure}

\begin{table}[!t]
\caption{The results of the MIAs.}
\label{tab:membershipattackresults}
\centering
    \renewcommand{\arraystretch}{1.3}
    \begin{tabular}{ccc}
    \hline
    \rowcolor{gray!20}                  & Resnet20 + CIFAR-10   & Resnet20 + CIFAR-100  \\ \hline
    Original                            & 95.62\%               & 94.91\%               \\
    \rowcolor{gray!20}Fully Retraining  & 0.00\%                & 0.00\%                \\
    Our Unlearning                      & 0.00\%                & 0.00\%                \\
    \hline
    \end{tabular}
\end{table}

\subsubsection{Evaluating Performance of Semantic Segmentation and Object Detection Tasks} In addition to multi-classification, we also evaluate our target unlearning scheme for the semantic segmentation scenario, where we use the pre-trained model $deeplabv3\_resnet50$ in PyTorch and set the unlearning target to \textit{person}. We set $\sigma = 26$, and $\delta=0.09$ to construct the essential graph and don't execute the unlearning process in the last two convolutional layers. We also evaluate our target unlearning scheme in an object detection scenario, where we choose $yolov5s$ as our model, and set $\sigma = 18$, $\delta=0.05$, unlearning target to \textit{person}. Experimental results are shown in Figure~\ref{fig:semanticsegmentation} and Figure~\ref{fig:objectdetection}, respectively.

\begin{figure*}[!t]
    \centering
    \subfloat[Grad-CAM of for Tiger Cat~(1)\centering]
    {\includegraphics[width=0.15\textwidth]{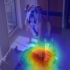}
    \label{fig_importantfeaturemaps:resnet18_GradCAM_cat_one}}
    \hfil
    \subfloat[Important weighted feature maps for Tiger Cat~(1)\centering]
    {\includegraphics[width=0.15\textwidth]{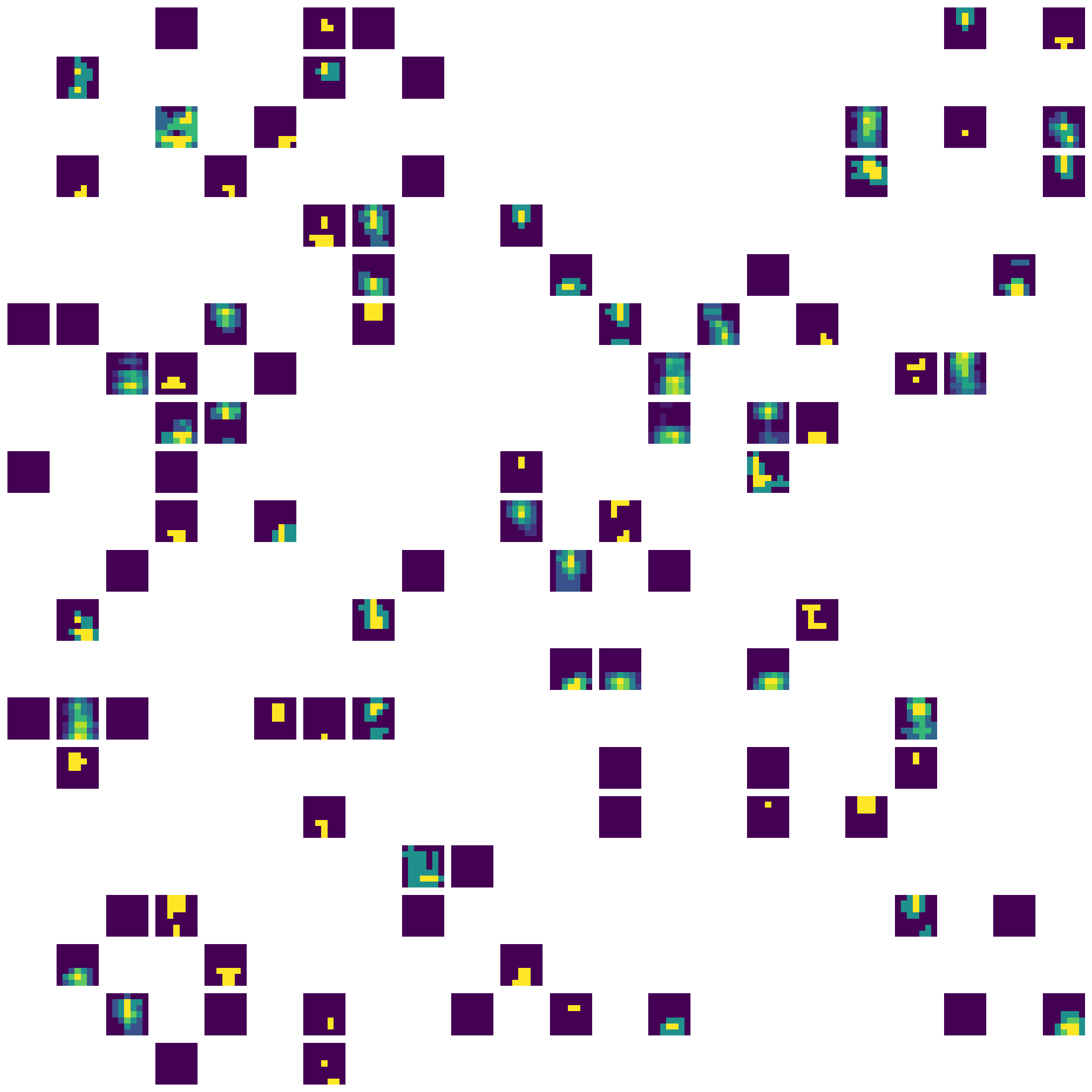}
    \label{fig_importantfeaturemaps:resnet18_GradCAM_cat_one_important}}
    \hfil
    \subfloat[Non-Important weighted feature maps for Tiger Cat~(1)\centering]
    {\includegraphics[width=0.15\textwidth]{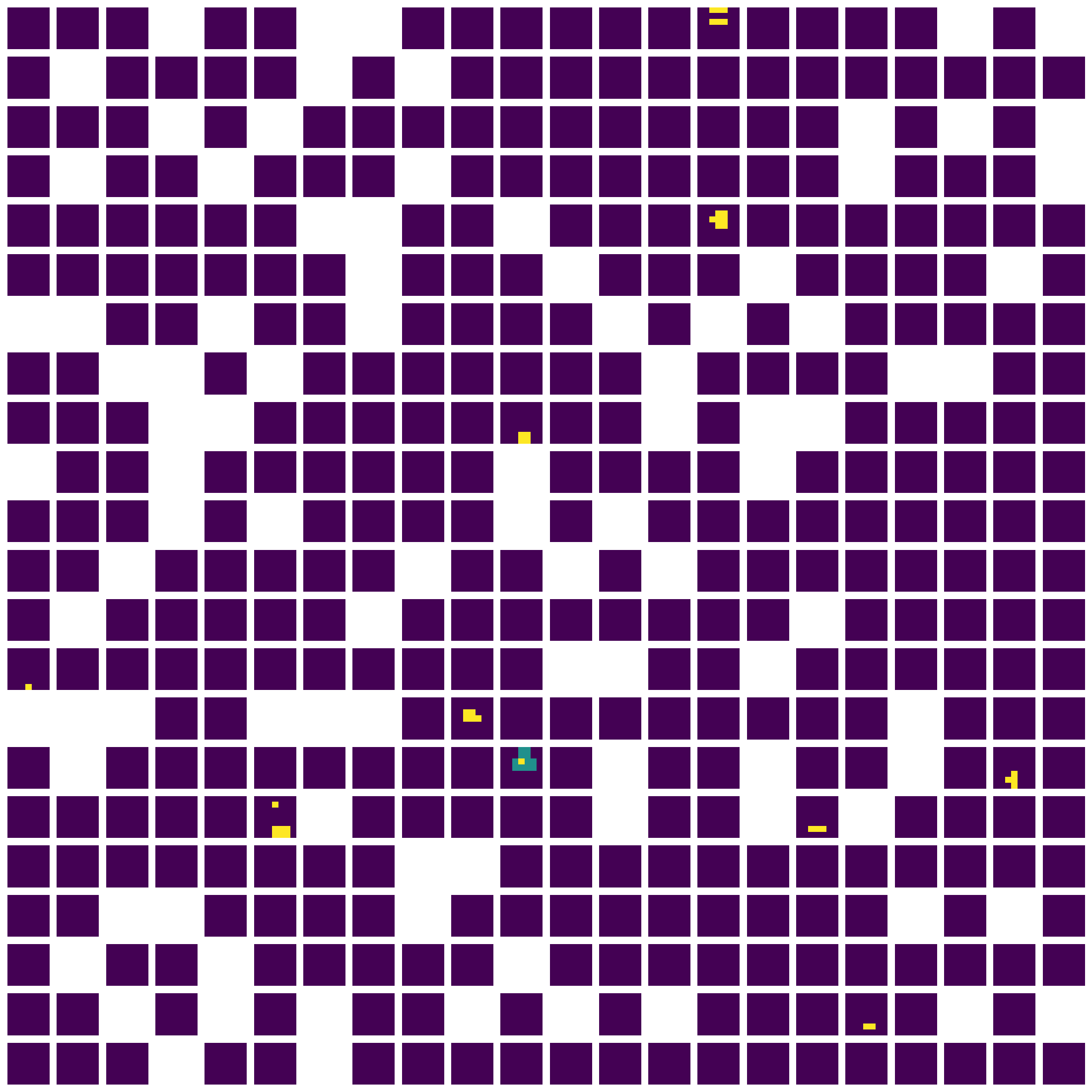}
    \label{fig_importantfeaturemaps:resnet18_GradCAM_cat_one_unimportant}}
    \subfloat[Grad-CAM of for Tiger Cat~(2)\centering]{\includegraphics[width=0.15\textwidth]{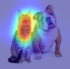}
    \label{fig_importantfeaturemaps:resnet18_GradCAM_cat_two}}
    \hfil
    \subfloat[Important weighted feature maps for Tiger Cat~(1)\centering]{\includegraphics[width=0.15\textwidth]{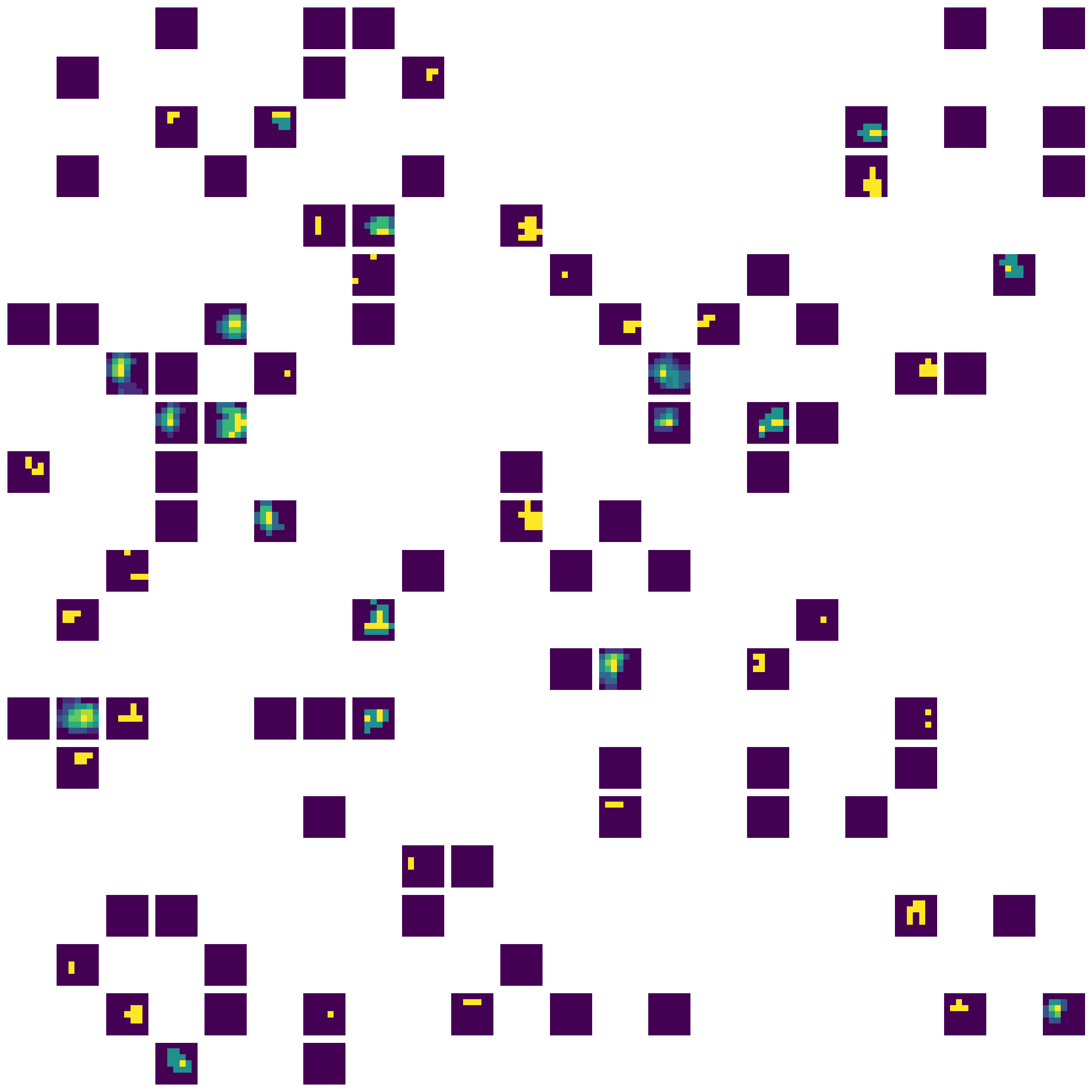}
    \label{fig_importantfeaturemaps:resnet18_GradCAM_cat_two_important}}
    \hfil
    \subfloat[Non-Important weighted feature maps for Tiger Cat~(1)\centering]{\includegraphics[width=0.15\textwidth]{imgs/showparameters/resnet18_GradCAM_cat_one_unimportant.png}
    \label{fig_importantfeaturemaps:resnet18_GradCAM_cat_two_unimportant}}
    \\
    \subfloat[Grad-CAM of for Boxer~(1)\centering]
    {\includegraphics[width=0.15\textwidth]{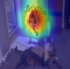}
    \label{fig_importantfeaturemaps:resnet18_GradCAM_dog_one}}
    \hfil
    \subfloat[Important weighted feature maps for Boxer~(1)\centering]
    {\includegraphics[width=0.15\textwidth]{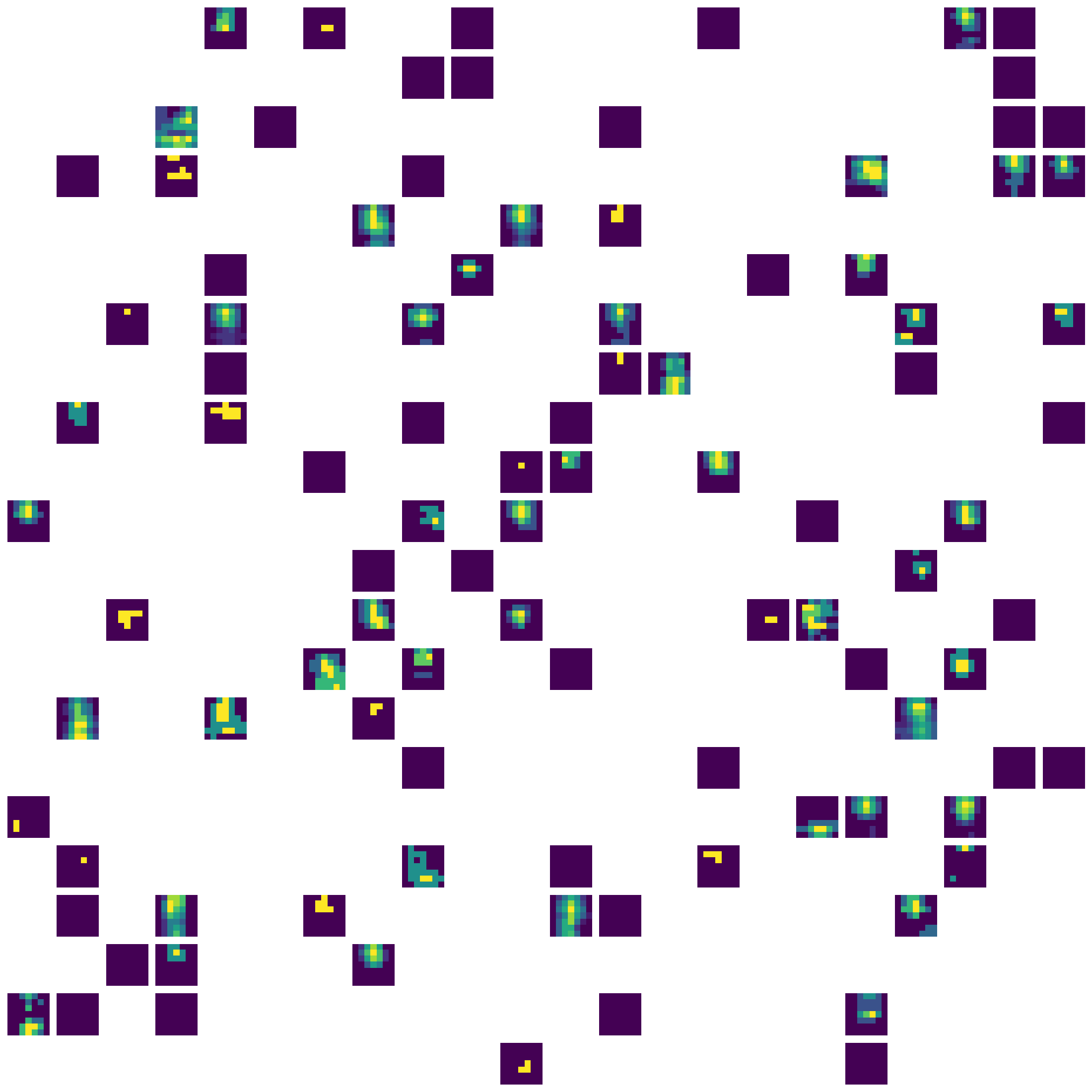}
    \label{fig_importantfeaturemaps:resnet18_GradCAM_dog_one_important}}
    \hfil
    \subfloat[Non-Important weighted feature maps for Boxer~(1)\centering]
    {\includegraphics[width=0.15\textwidth]{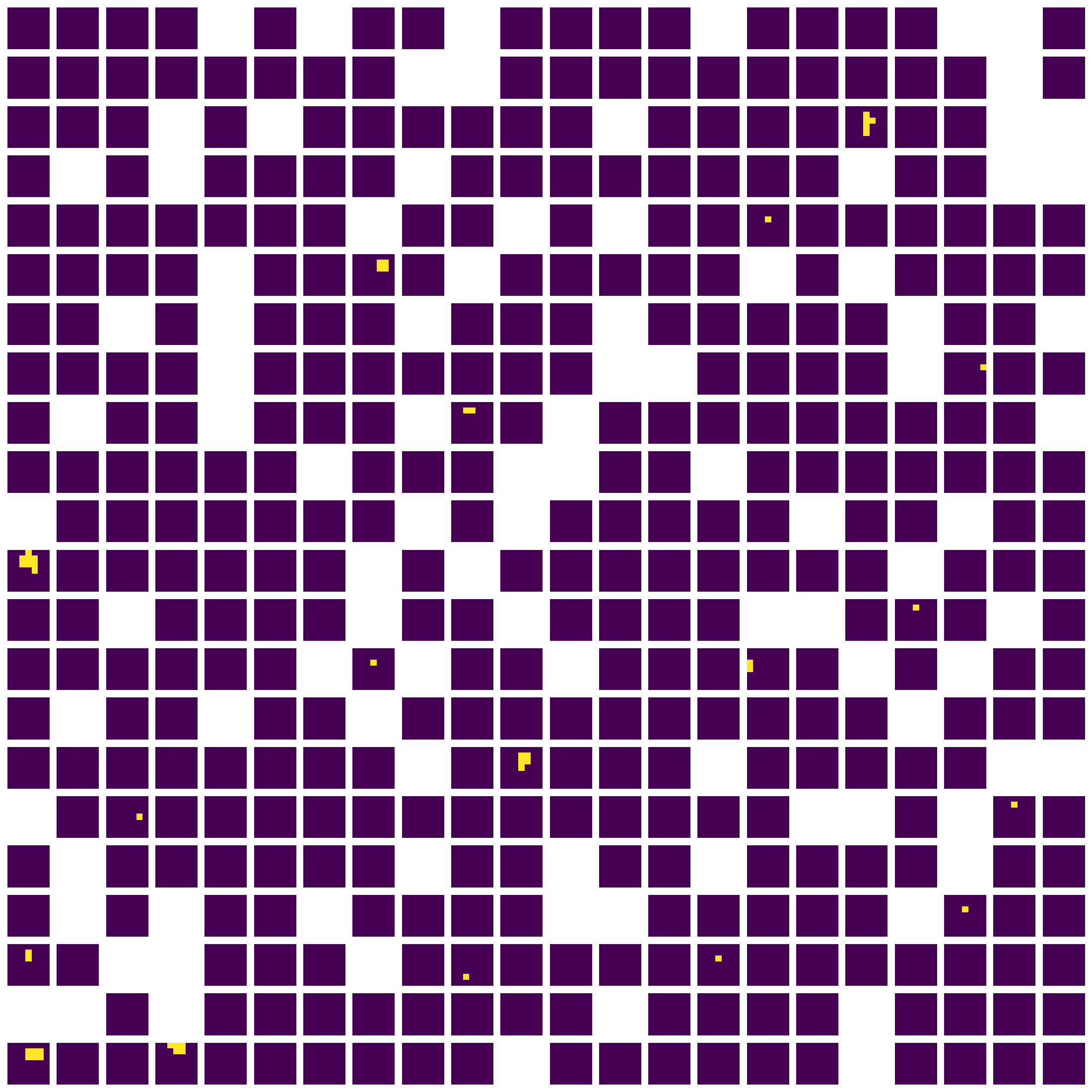}
    \label{fig_importantfeaturemaps:resnet18_GradCAM_dog_one_unimportant}}
     \subfloat[Grad-CAM of for Boxer~(2)\centering]{\includegraphics[width=0.15\textwidth]{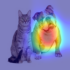}%
    \label{fig_importantfeaturemaps:resnet18_GradCAM_dog_two}}
    \hfil
    \subfloat[Important weighted feature maps for Boxer~(2)\centering]
    {\includegraphics[width=0.15\textwidth]{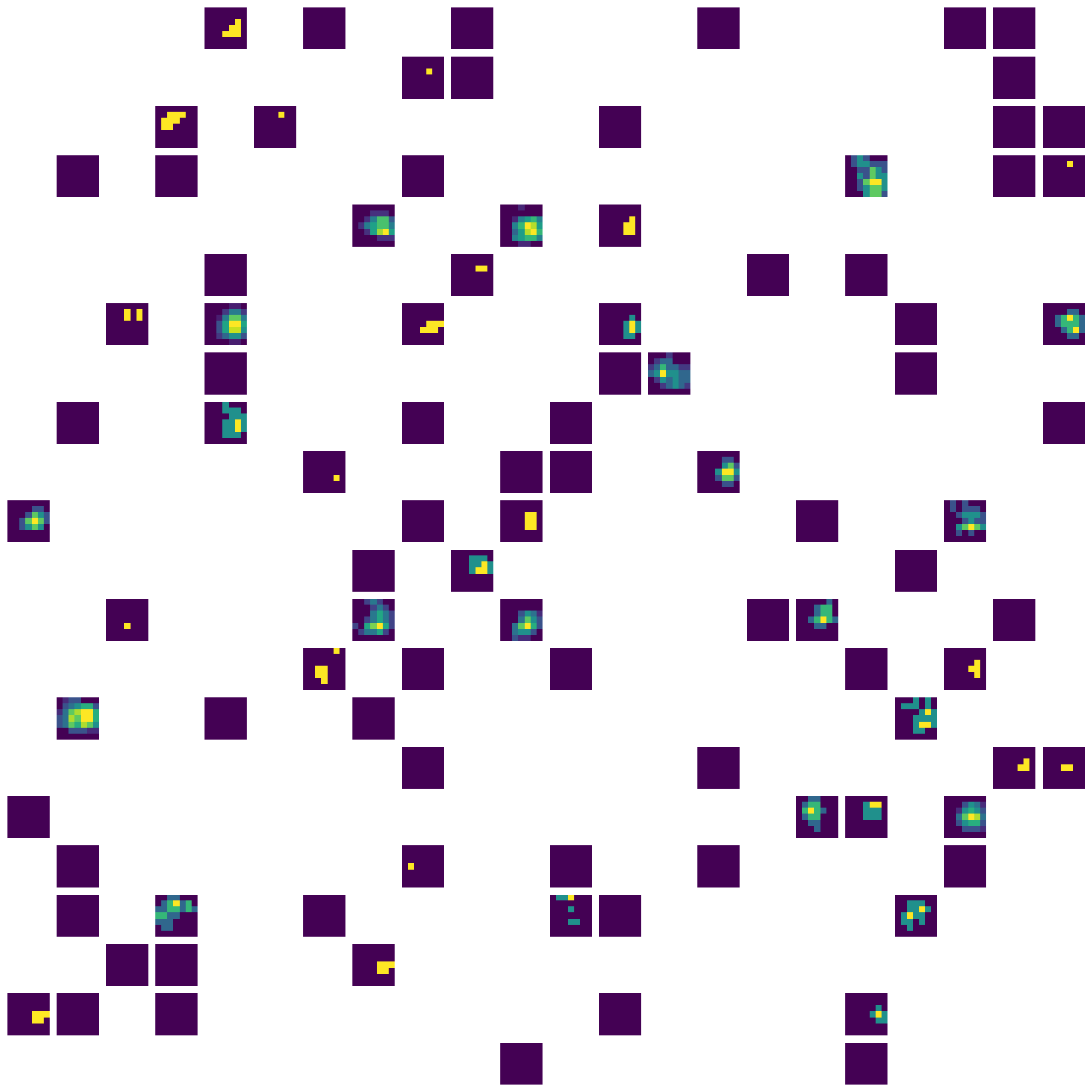}
    \label{fig_importantfeaturemaps:resnet18_GradCAM_dog_two_important}}
    \hfil
    \subfloat[Non-Important weighted feature maps for Boxer~(2) \centering]
    {\includegraphics[width=0.15\textwidth]{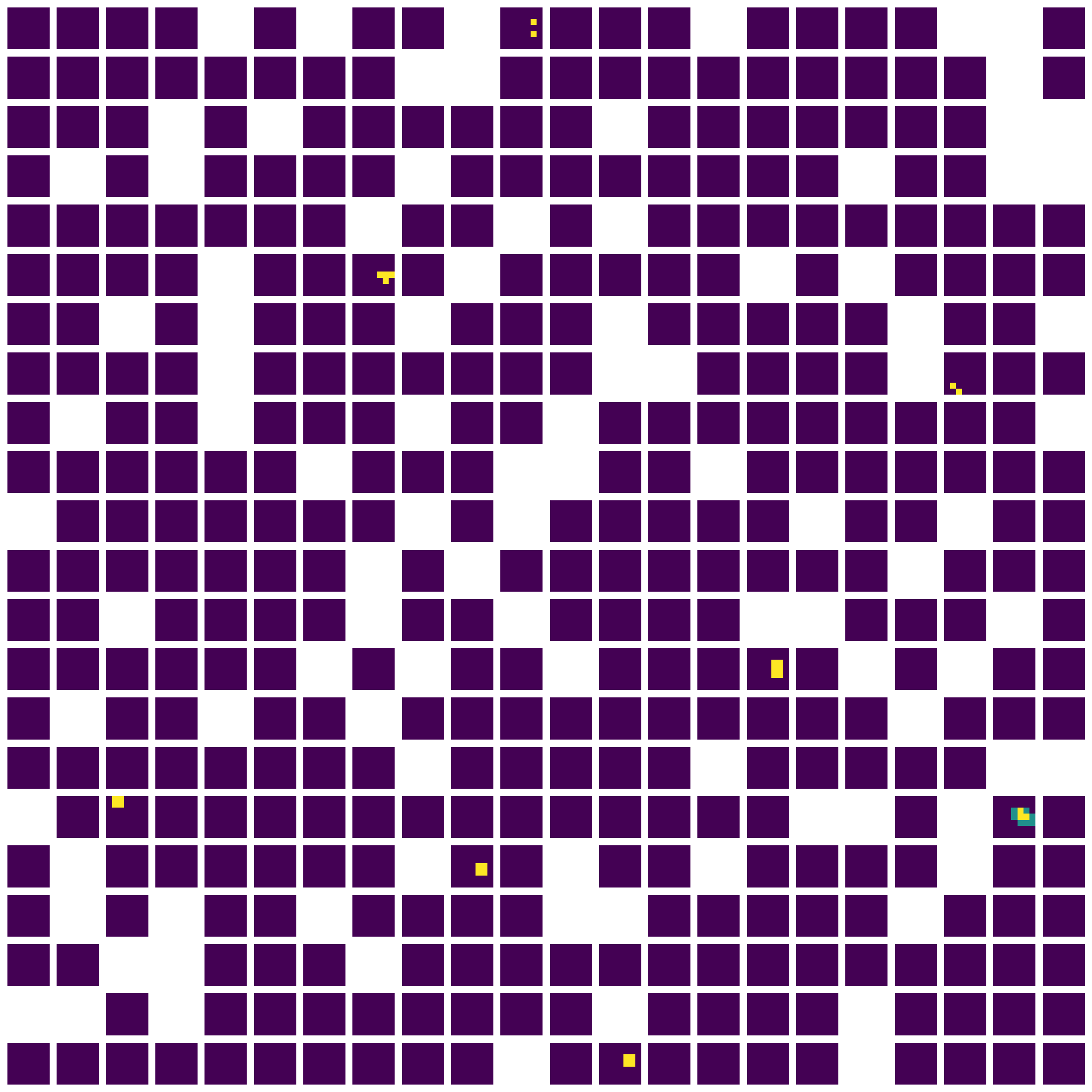}
    \label{fig_importantfeaturemaps:resnet18_GradCAM_dog_two_unimportant}}
    \caption{The results of Grad-CAM and the distribution of important and non-important feature maps toward different targets in ImageNet ILSVRC2012. All Grad-CAM results are calculated based on the last layer \textit{layer4.1.conv2} in ResNet18. For convenience, we only consider the first $22 \times 22$ feature maps, where the total number of the last layer in ResNet18 is $512$.}
    \label{fig:importantfeaturemaps}
\end{figure*}

\textbf{Results.} For semantic segmentation in Figure~\ref{fig:semanticsegmentation}, it can be seen that it is difficult for the semantic segmentation model to precisely recognize the person after the unlearning process. On the other hand, for bus, the unlearning operation essentially has no influence. For two results of object detection in Figure~\ref{fig:objectdetection}, similar results can be found. The object detection model can accurately detect objects such as horses, ovens, etc., while for person, it does not detect them.

\textbf{Summary.} The above experiments show that our scheme can achieve unlearning targets in different scenarios and can also effectively maintain model performance for all remaining tasks. In addition, it is more efficient since we achieve unlearning purposes based on only pruning the most critical parameters. For example, our scheme for multi-classification tasks based on CelebA only requires a few seconds, while the retraining process, usually takes multiple hours.

\subsubsection{Evaluating Unlearning based on Model Inversion}  To further analyze the effectiveness of our scheme, we construct the following experiment based on model inversion attack~\cite{DBLP:conf/ccs/FredriksonJR15} and membership inference attacks~\cite{DBLP:conf/csfw/YeomGFJ18}. Since those two schemes are typically used for instance-level evaluation, we consider a classification model where each class is treated as a distinct target. We implemented the model inversion attack as described in~\cite{DBLP:conf/ccs/FredriksonJR15}. We first train model ResNet18 based on the MNIST dataset with epoch = 50, batch size = 128 and learning rate = 0.1. We select the unlearning target as class 3 and consider other class data as the remaining data. After the training process, we select the critical parameters for unlearning class $3$ and set the $\sigma=5$, $\delta = 0.6$, then prune those critical parameters, followed by one epoch to recover model performance. Figure~\ref{fig:unlearningdataaccuracy} illustrates the results of model inversion attacks against models affected by different unlearning methods.

\textbf{Results.} As shown in Figure~\ref{fig:unlearningdataaccuracy}, fully retraining and our unlearning scheme produce dark and jumbled images since the model inversion attack has relatively little gradient information to rely on. As expected, pruning the critical parameters and fine-tuning prevent the inference of any meaningful information about the unlearning target. This suggests that our unlearning scheme almost removes the information about the unlearning target and eliminates the potential of inferring useful information via model inversion attacks.

\subsubsection{Evaluating Unlearning based on Membership Inference Attack} Additionally, we use MIAs in paper~\cite{DBLP:conf/csfw/YeomGFJ18} to evaluate whether the unlearning instances are still identifiable in the training dataset. We set the number of shadow models as 20 and the training epoch of the shadow model as 10, batch size = 64. The attack model is a fully connected network with two hidden linear layers of width 256 and 128, respectively, with ReLU activation functions and a sigmoid output layer. We evaluate our unlearning scheme with two settings, CIAFR-10 + ResNet20 and CIFAR-100 + ResNet20, and set unlearning class = 3. In the case of the CIFAR100 dataset, after sorting the model outputs, we choose the top 10 outputs as inputs for the attack model. We equally divide the training dataset into two subsets to generate the dataset based on shadow models and then train the attack model based on the output of those shadow models. After that, we set $\delta = 0.2$, $\sigma = 5$ and $\delta = 0.08$, $\sigma = 5$ to construct balanced essential graphs, respectively, and prune the critical parameters to unlearn target data. We set epoch = 5 to recover the model performance before MIAs. Table~\ref{tab:membershipattackresults} shows the results of our evaluation.

As shown from Table~\ref{tab:membershipattackresults}, All MIAs have a high success rate for all original models; i.e., they can successfully derive the training dataset containing unlearning targets. However, the success rate of all MIAs is lower for all other approaches, indicating that MIAs cannot determine the existence of the unlearning targets after the unlearning process; this suggests that the influence of the unlearning target has been effectively removed from the unlearned model.

\textbf{Summary.} The above experiments show that our scheme can effectively obtain critical parameters containing information about the unlearning targets, and reasonable pruning can reduce the probability of an attacker obtaining confidential information about those targets.

\subsection{Feasibility Analysis}
Our unlearning solution is based on the fact that the intermediate weights $\alpha_{l}^k$ in Equation~\ref{equation:2} represent the importance of the corresponding channel's parameter to a target~\cite{DBLP:conf/cvpr/WangWDYZDMH20,DBLP:journals/tip/JiangZHCW21}. To demonstrate the feasibility of our approach, we do the following experiments to verify that (1). given one instance, the distribution of intermediate weights is similar for all instances with the same target data; and (2). given the important parameters in a layer of a target, a reasonable disturbance to those parameters will destroy the performance of this target data. Here, the distribution of intermediate weights means the set of important parameters and non-important parameters for a specific target. Object (1) ensures that all instances with the same target will be influenced by the same important parameters, which guarantees that the subsequent disturbances will influence the same parameters to unlearn the target. Object (2) ensures the feasibility of achieving unlearning based on the selected important parameters.

\begin{figure}[!t]
    \centering
    \subfloat[Alexnet~(Unlearning Target)\label{fig:importanceanalysisalexnetunlearning}]{\includegraphics[width=0.25\textwidth]{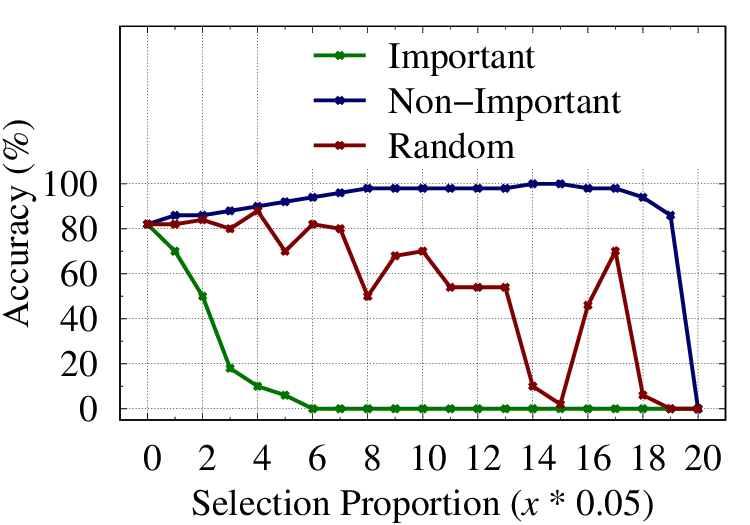}}
    \subfloat[Alexnet~(Remaining Data)\label{fig:importanceanalysisalexnetremaining}]
    {\includegraphics[width=0.25\textwidth]{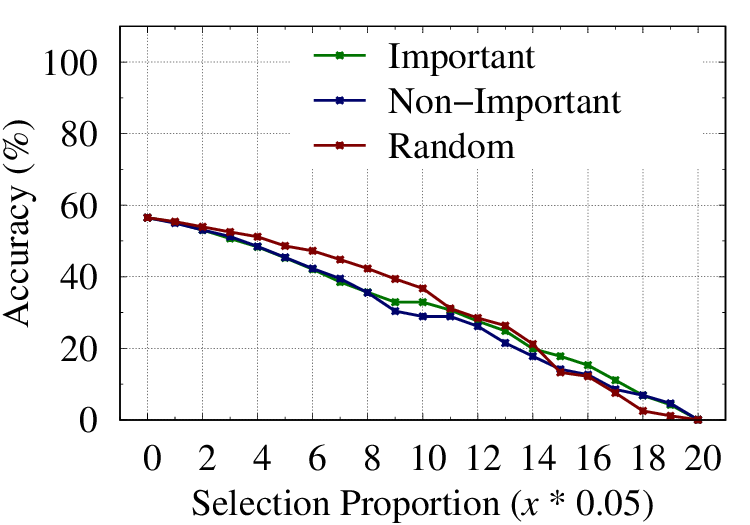}}\\
    \subfloat[VGG13~(Unlearning Target)\label{fig:importanceanalysisvgg13unlearning}]{\includegraphics[width=0.25\textwidth]{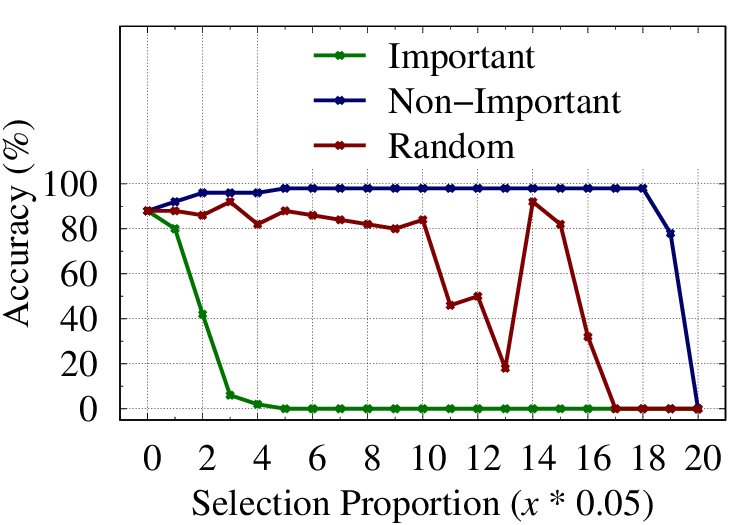}}
    \subfloat[VGG13~(Remaining Data)\label{fig:importanceanalysisvgg13remaining}]
    {\includegraphics[width=0.25\textwidth]{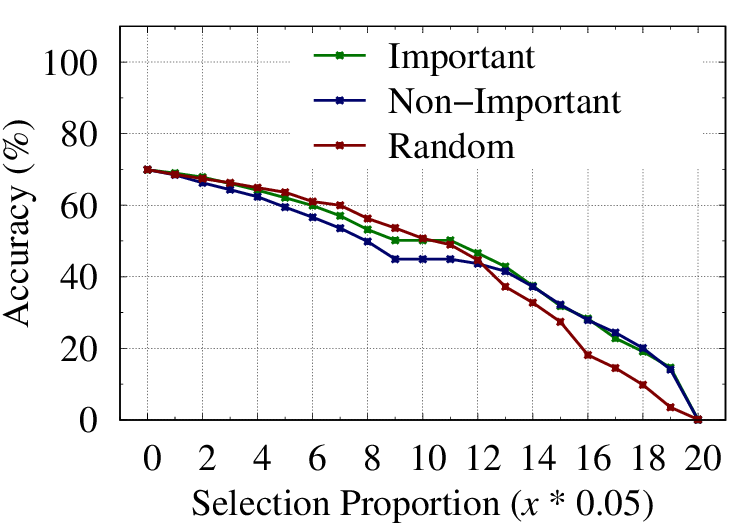}}
    \caption{The model accuracy after pruning different types of parameters.}
    \label{fig:importanceanalysis}
\end{figure}

\begin{figure}[!t]
    \centering
    \subfloat[Essential Graph without Balance~\textbf{Graph}]{\includegraphics[width=0.45\textwidth]{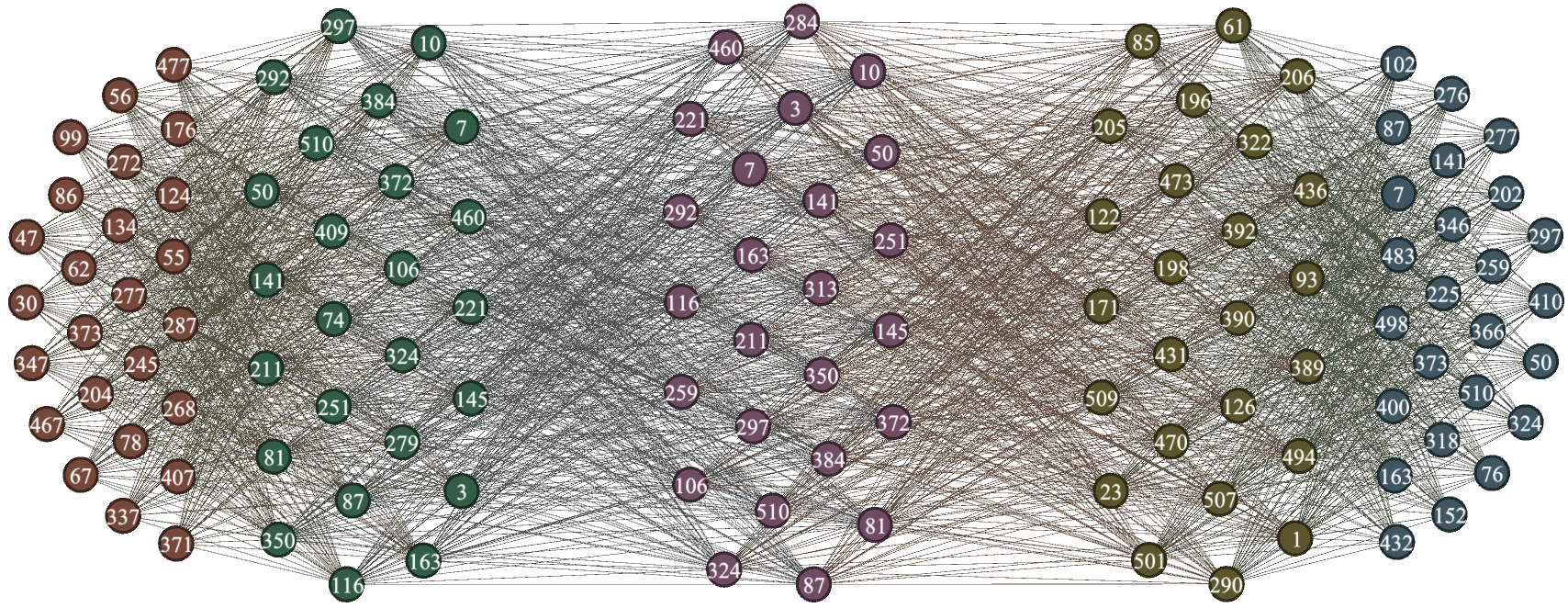}
    \label{fig:originalgraph}}\\
    \subfloat[Essential Graph with Balance~(\textbf{Graph~($\beta$)})]{\includegraphics[width=0.38\textwidth]{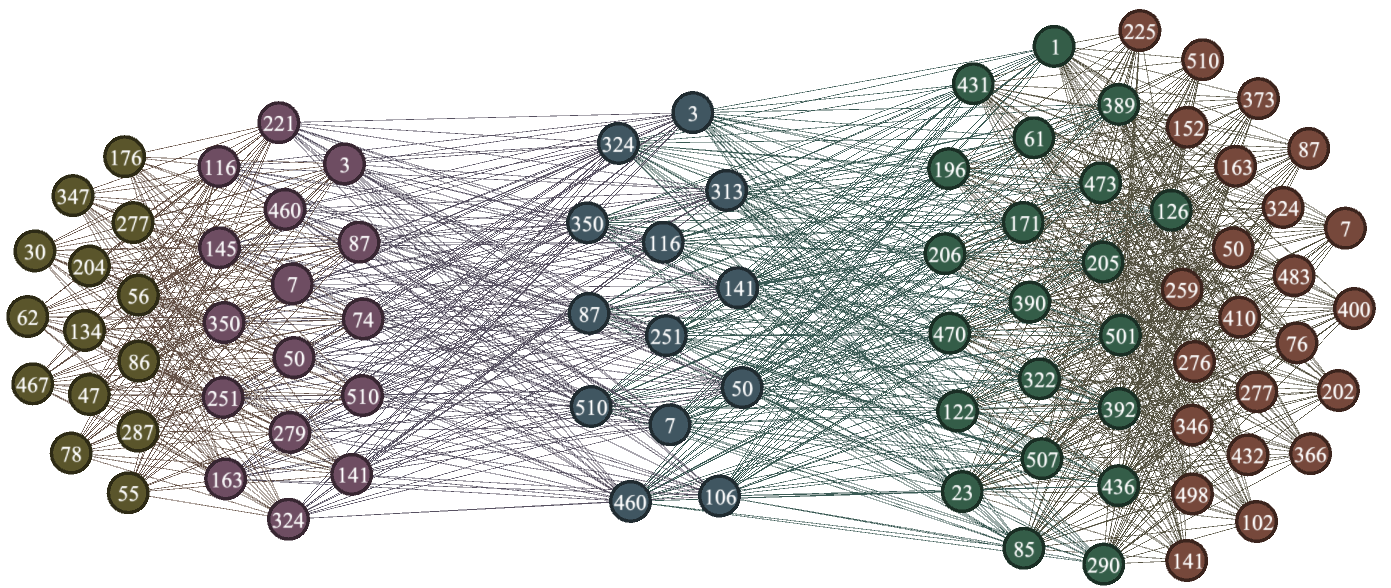}
    \label{fig:seletedgraph}}
    \caption{Different essential graphs.}
    \label{fig:graph}
\end{figure}

\subsubsection{Identity Analysis} To analyze if the distribution of intermediate weight is similar for all instances containing the same targets, we show the distribution of weighted feature maps of the final convolutional layer using the hook technique. In particular, we separately random select two instances from ImageNet ILSVRC2012, and compute the Grad-CAM toward the final layer \textit{layer4.1.conv2} in ResNet18 model using pre-trained weights in PyTorch. During the calculation of GradCAM, we also record the intermediate weights $\alpha^k$ of each feature map, and then calculate the results of $\alpha^k \times f^{k}$. We divide the top 20\% of the results of $\alpha^k \times f^{k}$ based on the value of $\alpha^k$ as important weighted feature maps and take the remaining 80\% as non-important weighted feature maps. We separately plot the above two types of weighted feature maps and maintain the position of each feature map. Ideally, if the index of selected important weighted feature maps of two instances are similar, it means that the distribution of intermediate weights $\alpha^k$ is similar.

\textbf{Results.} Figure~\ref{fig:importantfeaturemaps} illustrate the results of our experiment. Figure~\ref{fig_importantfeaturemaps:resnet18_GradCAM_cat_one} and Figure~\ref{fig_importantfeaturemaps:resnet18_GradCAM_cat_two} represent the Grad-CAM of \textit{Tiger Cat}, while Figure~\ref{fig_importantfeaturemaps:resnet18_GradCAM_dog_one} and Figure~\ref{fig_importantfeaturemaps:resnet18_GradCAM_dog_two} show the Grad-CAM of \textit{Boxer}. The other subplots show the index of importance and non-importance weighted feature maps. We can see from the Figure~\ref{fig_importantfeaturemaps:resnet18_GradCAM_cat_one_important} and Figure~\ref{fig_importantfeaturemaps:resnet18_GradCAM_cat_two_important} that for \textit{Tiger Cat}~(1) and \textit{Tiger Cat}~(2), the index of the weighted important feature maps is almost same, and from the Figure~\ref{fig_importantfeaturemaps:resnet18_GradCAM_cat_one_unimportant} and Figure~\ref{fig_importantfeaturemaps:resnet18_GradCAM_cat_two_unimportant}, the index of non-important weighted feature maps is also nearly identical. That is to say, the important parameters within one layer are almost the same for those two different instances with the same target data. This indirectly suggests that the distribution of those two intermediate weights is almost identical. The same results can be derived from Figure~\ref{fig_importantfeaturemaps:resnet18_GradCAM_dog_one_important} and Figure~\ref{fig_importantfeaturemaps:resnet18_GradCAM_dog_two_important}. In addition, by comparing Figure~\ref{fig_importantfeaturemaps:resnet18_GradCAM_cat_one_important} and Figure~\ref{fig_importantfeaturemaps:resnet18_GradCAM_dog_one_important}~(Figure~\ref{fig_importantfeaturemaps:resnet18_GradCAM_cat_two_important} and Figure~\ref{fig_importantfeaturemaps:resnet18_GradCAM_dog_two_important}), we can also find that the distribution of intermediate weights generated from two instances with various target labels are different. At the same time, there is some overlap between Figure~\ref{fig_importantfeaturemaps:resnet18_GradCAM_cat_one_important} and Figure~\ref{fig_importantfeaturemaps:resnet18_GradCAM_dog_one_important}~(Figure~\ref{fig_importantfeaturemaps:resnet18_GradCAM_cat_two_important} and Figure~\ref{fig_importantfeaturemaps:resnet18_GradCAM_dog_two_important}), this suggests that different targets may use the same parameters to analyze instances and give the prediction of those targets. Therefore, it is necessary to consider the model performance of the remaining data when executing the unlearning process.

\begin{table*}[]
\caption{Experimental settings in evaluating the essential graph with balance.}
\label{tab:experimentsetting_balance}
\centering
    \renewcommand{\arraystretch}{1.3}
    \begin{tabular}{ccccc}
    \hline
    Different settings &  Original targets        & Target that need to be unlearned    & $\delta$  & $\sigma$ \\ \hline
    I         &Smiling, No Beard, Eyeglasses               & No Beard                     & 0.1       & 6   \\
    II        &Mouth Slightly Open, No Beard, Eyeglasses   & No Beard                     & 0.25      & 7     \\
    III       &Mouth Slightly Open, No Beard, Wearing Hat  & Mouth Slightly Open          & 0.18      & 7     \\
    IV        &Smiling, No Beard, Wearing Hat              & Smiling                      & 0.3       & 7    \\\hline
    \end{tabular}
\end{table*}

\subsubsection{Important Parameter Analysis} To verify if a reasonable disturbance to those important parameters will destroy the performance of target data. We further measure the accuracy of one unlearning target and the remaining data after the unlearning process, respectively. Specifically, we first select two different models with pre-trained weights in PyTorch, including AlexNet, and VGG13, to calculate the intermediate weights based on Grad-CAM in the last layer toward the unlearning target \textit{Tinca} in ImageNet ILSVRC2012. Then, we sort those intermediate weights from large to small and select the top large weights for different proportions. We also compare two other weight selection schemes, including random and non-important selection, where for non-important selection, we select the smallest intermediate weights. After that, we prune model parameters corresponding to the index of these selected weights and record the accuracy. Figure~\ref{fig:importanceanalysis} illustrates the results of our experiment.

\textbf{Results.} In Figure~\ref{fig:importanceanalysis}, Figure~\ref{fig:importanceanalysisalexnetunlearning} and Figure~\ref{fig:importanceanalysisalexnetremaining} show the accuracy of target that need to be unlearned and remaining data of AlexNet, while Figure~\ref{fig:importanceanalysisvgg13unlearning} and Figure~\ref{fig:importanceanalysisvgg13remaining} show the results of VGG13. The y-axis denotes the accuracy, while the x-axis represents the different proportions of the selected intermediate weights. The x-axis also indicates how much the proportion of parameters we prune. From all results, we find that as we prune more model parameters, the accuracy of unlearning target and remaining data gradually decreases. This is because as more parameters are pruned, the model contains progressively less information about those data, which will lead to a decrease in accuracy. It is worth noting that in the front part of the x-axis for the important selection scheme, that is $x \in (0:6]$, the accuracy of unlearning target decreases significantly, while in the remaining part of the x-axis, the accuracy remains almost zero. This indicates that some model parameters are only associated with unlearning targets. For the non-important or random selection scheme, the performance for the unlearning target is almost unchanged in the front part of the x-axis, which indicates that these two methods cannot select the parameters that affect the unlearning target.

\textbf{Summary.} The above experiments have shown that the distribution of the most influential parameters for the same target is similar. Reasonable pruning to those selected influential parameters will affect the model performance for those target data without affecting the remaining data, which provides an effective unlearning solution.

\begin{figure}
    \centering
    \subfloat[Setting I]{\includegraphics[width=0.25\textwidth]{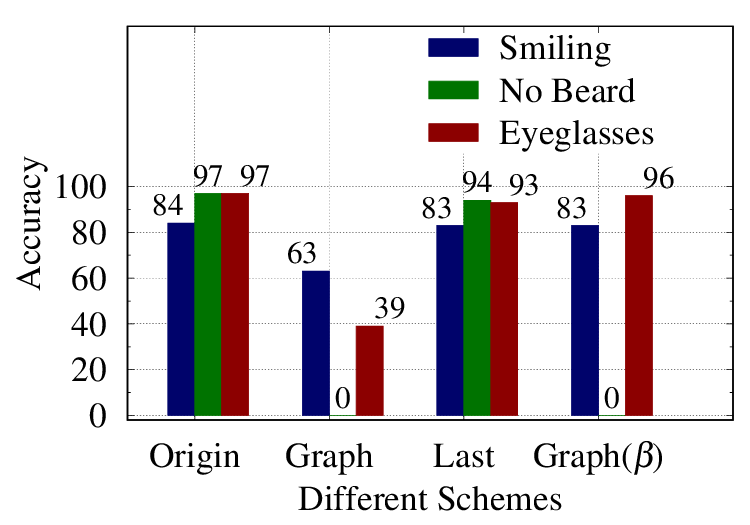}
    \label{fig:resnet20cifar10}}
    \subfloat[Setting II]{\includegraphics[width=0.25\textwidth]{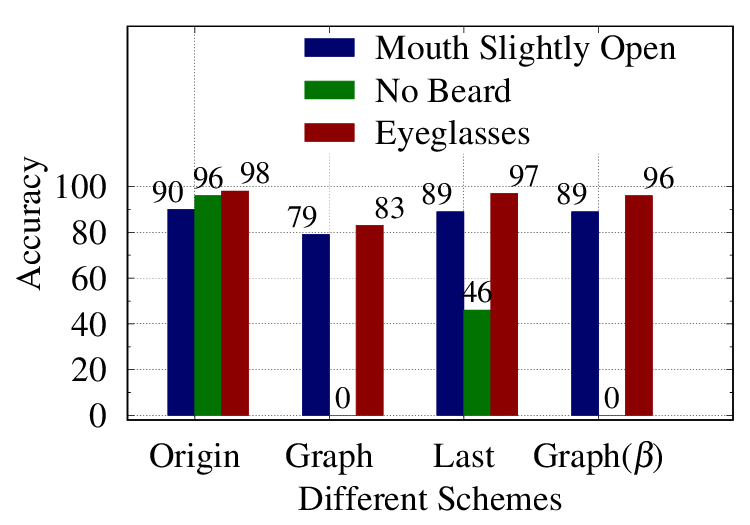}
    \label{fig:resnet20cifar100}}\\
    \subfloat[Setting III]{\includegraphics[width=0.25\textwidth]{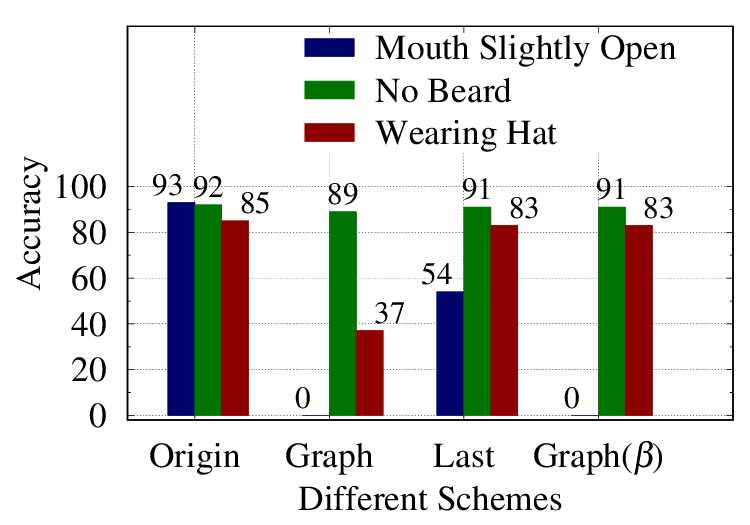}
    \label{fig:imagenetresnet18}}
    \subfloat[Setting IV]{\includegraphics[width=0.25\textwidth]{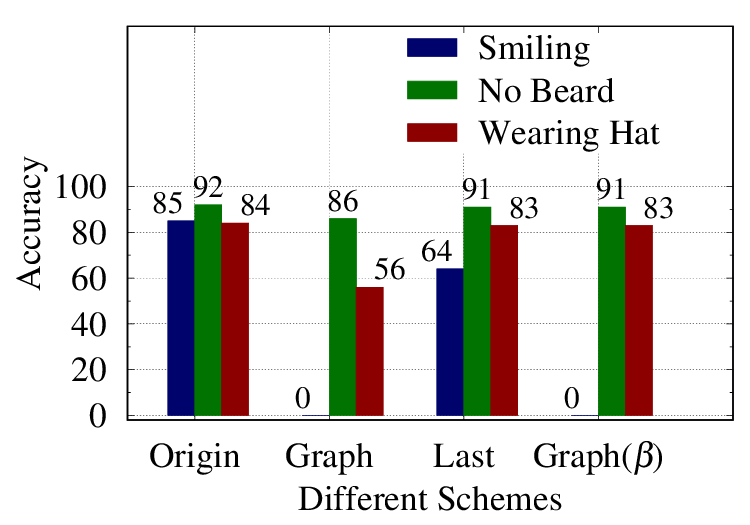}
    \label{fig:imagenetvgg13}}\\
    \caption{The results of evaluating balance graph.}
    \label{fig:accracyresults}
\end{figure}

\subsubsection{Essential and Balance Graph Analysis} Some of the parameters that affect the unlearning target may also be important for the remaining data. Directly removing information contained in those parameters that are important for the unlearning target may simultaneously affect the model performance for the remaining data. In Section~\ref{subsec:efficientunlearningprocess}, we construct the graph to balance unlearning effectiveness and model performance. To evaluate the validity of the balanced essential graph, we construct the following experiments. We use the pre-trained model ResNet18 in PyTorch and set unlearning target label = 0, $\sigma = 5$, and set $\delta = 0.05$ to construct the essential graph based on the unlearning target, which is denoted as \textbf{Graph}. We also construct the balanced essential graph while simultaneously considering the remaining data and unlearning data, which we denote as \textbf{Graph~($\beta$)}. Figure~\ref{fig:graph} shows the results of our experiment, where nodes with the same color denote each layer of the model, and nodes in each partition represent the corresponding channels. For example, the rightmost partition in Figure~\ref{fig:originalgraph} represents the last layer in ResNet18, while the node $7$ in this partition indicates the corresponding channel $7$ in the last layer.

\textbf{Results.} In Figure~\ref{fig:graph}, Figure~\ref{fig:originalgraph} shows the essential graph constructed by unlearning target data only, while Figure~\ref{fig:seletedgraph} shows the essential graph constructed by balancing the impact of the remaining data. There are two main differences between these two graphs. First, some nodes exist in Figure~\ref{fig:originalgraph}  will disappear in the corresponding partitions in Figure~\ref{fig:seletedgraph}. For example, the node $297$ in the last partition in Figure~\ref{fig:originalgraph}. In the last partition in Figure~\ref{fig:seletedgraph}, node $297$ doesn't exist. This indicates that parameters in one layer of the model that are important for the unlearning target may also be important for the remaining data. During the construction of the essential graph with balancing the impact of the remaining data, those parameters that are also important for the remaining data will be discharged. Second, the number of nodes owned by each color partition in Figure~\ref{fig:originalgraph} is similar, while in Figure~\ref{fig:seletedgraph}, the number of nodes in each partition is different, with fewer nodes on the left. The reason for this is that the lower layers of the model are usually used to extract features, while the parameters at the higher levels of the model are used to distinguish between different targets. Lower-level features always have a higher probability of being used by the remaining data. When constructing the essential graph in Figure~\ref{fig:seletedgraph}, since we also consider the performance of the unlearned model for the remaining data, it will select a smaller number of important nodes to avoid degrading the performance of the remaining data.

\begin{figure}[!t]
    \centering
    \subfloat[$\sigma = 3$]{\includegraphics[width=0.25\textwidth]{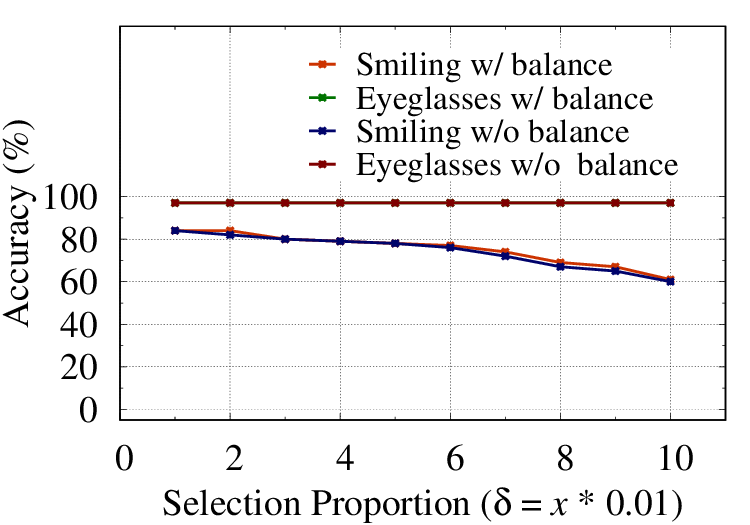}
    \label{fig:hypyer_3}}
    \subfloat[$\sigma = 4$]{\includegraphics[width=0.25\textwidth]{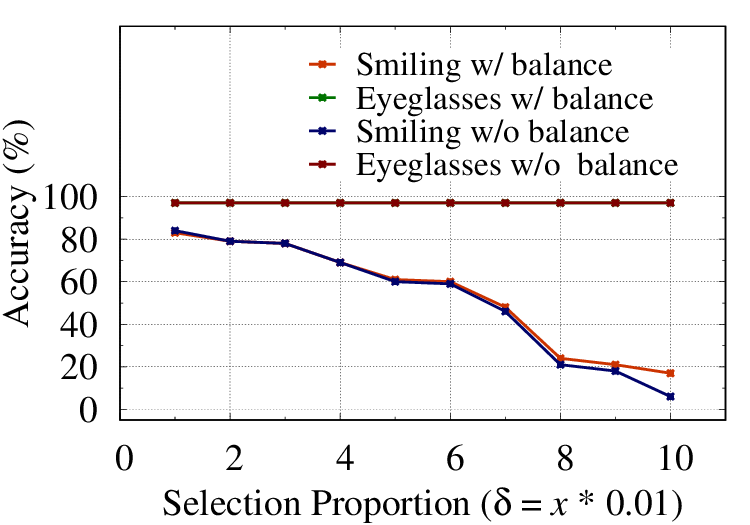}
    \label{fig:hypyer_4}}\\
    \subfloat[$\sigma = 5$]{\includegraphics[width=0.25\textwidth]{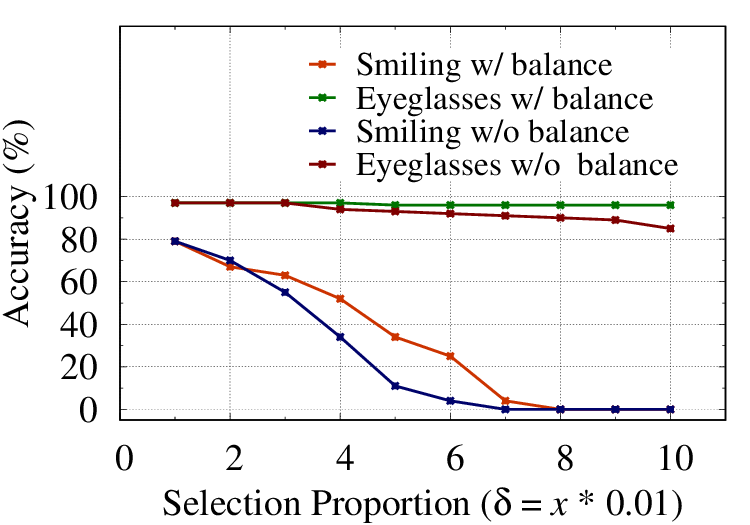}
    \label{fig:hypyer_5}}
     \subfloat[$\sigma = 6$]{\includegraphics[width=0.25\textwidth]{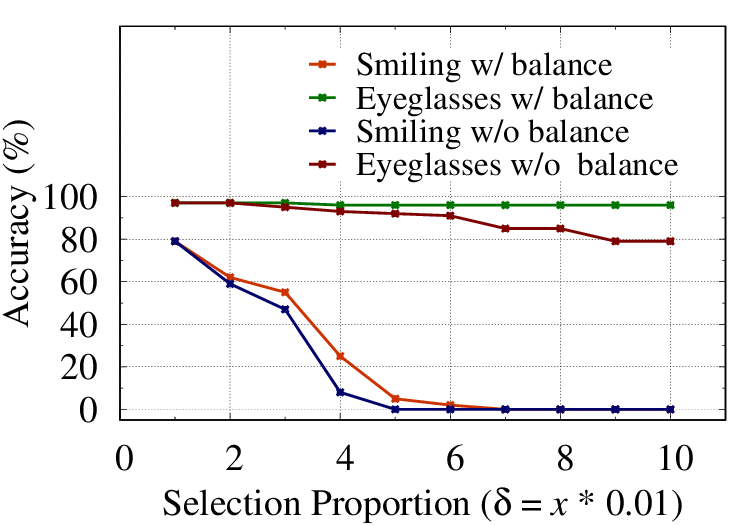}
    \label{fig:hypyer_6}}
    \caption{The effect of hyper-parameters.}
    \label{fig:hypyer}
\end{figure}

\subsubsection{Evaluating Balance Graph} In this Section, we evaluate whether the balanced essential graph can maintain the performance of the remaining data while ensuring the effectiveness of the unlearning. As shown in Table~\ref{tab:experimentsetting_balance}, we consider four different settings. For all original model training processes, we set epoch = 20, batch size = 128, and learning rate = 5e-06. We compare three unlearning schemes, including only executing the unlearning process in the last layer (\textbf{Last}), unlearning based on the essential graph without balance~(\textbf{Graph}) and unlearning based on the essential graph with balance~(\textbf{Graph~($\beta$)}). The results are shown in Figure~\ref{fig:accracyresults}.

\textbf{Results.} As can be seen in Figure~\ref{fig:accracyresults}, when only considering the unlearning process in the last layer~(\textbf{Last}), the performance of the unlearned model for targets that need to be unlearned hardly decreases to 0, indicating that the model still contains some information about those targets. For the unlearning scheme based on the graph without balance~(\textbf{Graph}), the accuracy of unlearning target data is completely reduced to $0$. However, it does not meet the requirements for the utility of the remaining data. While the parameters associated with the unlearning target are pruned, those parameters that are also important for other targets in the constructed essential graph are also pruned, which will destroy the performance of the remaining data. For \textbf{Graph~($\beta$)}, it can maintain the performance of the model after the unlearning while ensuring the effectiveness of the unlearning.

\textbf{Summary.} The above experiments demonstrate the effectiveness of the balanced essential graph. It is able to select the important parameters while controlling the distribution of the selected parameters as well, reducing model performance degradation due to over-selection of important parameters.

\subsection{The Effect of Hyper-Parameters}
\textbf{Setup.} The hyperparameter $\sigma$ denotes the number of layers that are considered to execute the unlearning process, while the $\delta$ determines how many parameters are selected when constructing the graph. To evaluate both hyperparameters, we set the original targets as smiling, no beard, and eyeglasses in CelebA, and we set the unlearning target as smiling. For the original model training processes, we set epoch = 20, batch size = 128, and learning rate = 5e-06. Then, we, respectively, choose different values of $\sigma$ and $\delta$ to construct the essential graph with balance and without balance. Then, we evaluate the accuracy of smiling and eyeglasses after pruning the most critical parameters. Figure~\ref{fig:hypyer} shows the results.

\textbf{Results.} As seen from all Figures, when keeping $\delta$ constant and increasing $\sigma$, the number of layers to be pruned will increase, which directly accelerates the unlearning process. Similar results can also be observed from each Figure when setting $\sigma$ to the same value and increasing $\delta$. It's worth noting that as $\delta$ increases, the accuracy of the remaining targets will decrease when constructing graphs without balance. For example, in Figure~\ref{fig:hypyer_5}, when setting  $\delta$ = 0.1 and $\sigma$ = 5, the model performance for eyeglasses decreases to $81\%$. On the other hand, even when selecting a larger number of $\delta$ and $\sigma$, the accuracy of the remaining targets does not decrease for all schemes with balance operations. This is because when constructing \textbf{Graph~($\beta$)}, we simultaneously consider the performance of the model for the remaining data, which can alleviate the performance decrease for the remaining data. Based on the above results, our unlearning scheme can achieve unlearning effectiveness while ensuring model usability even when both hyperparameters are set too high. Therefore, in general, we can opt for appropriately large hyperparameters to achieve effective unlearning purposes.

\section{Conclusion and Future Work}
\label{sec:conclusion}

In this paper, we have proposed a novel machine unlearning scheme that can selectively remove partial target information from the trained model, name \textit{target unlearning}. As a solution, we have defined the concept of target unlearning and illustrated the challenges of this unlearning problem. We also analyzed the most influential parameters of a model for the given target based on the explainable technique and proposed a pruning-based unlearning method to erase the information about the target. To balance the performance of the unlearned model, we constructed an essential graph to describe the relationship between all important parameters within the model, and simultaneously filter those important parameters that are also important for the remaining data. The experimental results demonstrate that under our scheme, the model can remove the impact of targets and ensure the accuracy of the remaining data in a quick and efficient manner.

Since the CAM-based techniques can only be applied to convolutional neural networks, which would lead to limiting our scheme to CNN models, in the future, we will explore ways to extend and/or modify the current method of evaluating channel important and develop new target unlearning schemes based on this revised evaluation. In addition, we plan to design a more powerful scheme that supports unlearning requests from Natural Language Processing~(NLP) or Generative Adversarial Networks~(GAN), combined with other technologies such as information theory.

\ifCLASSOPTIONcaptionsoff
  \newpage
\fi

\bibliographystyle{IEEEtran}
\bibliography{sampleBibFile}

\begin{IEEEbiography}[{\includegraphics[width=1in,height=1.25in,clip,keepaspectratio]{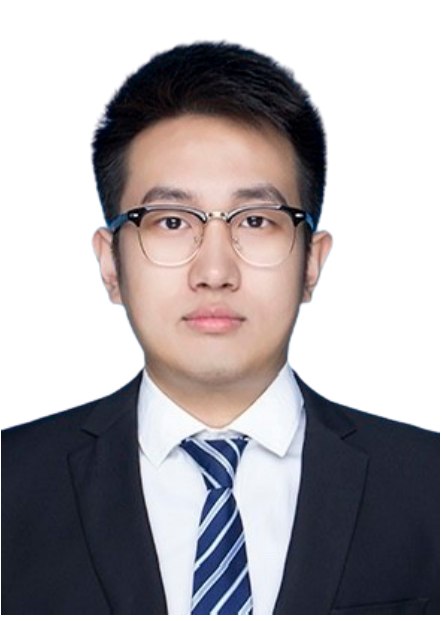}}]{Heng Xu} received his B.Eng. and M.Eng. degrees from China University of Geosciences, China, in 2018 and 2021, respectively. He is currently a PhD student at the University of Technology Sydney, Australia. His research interests are machine unlearning.
\end{IEEEbiography}

\begin{IEEEbiography}[{\includegraphics[width=1in,height=1.25in,clip,keepaspectratio]{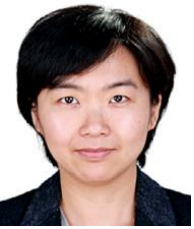}}]{Tianqing Zhu} received her BEng and MEng degrees from Wuhan University, China, in 2000 and 2004, respectively, and a PhD degree from Deakin University in Computer Science, Australia, in 2014. Dr. Tianqing Zhu is currently a professor in the faculty of data science at the City University of Macau. Before that, she was a lecturer at the School of Information Technology, Deakin University, Australia, and an associate professor at the University of Technology Sydney, Australia. Her research interests include privacy-preserving and AI security. 
\end{IEEEbiography}

\begin{IEEEbiography}[{\includegraphics[width=1in,height=1.25in,clip,keepaspectratio]{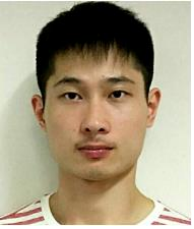}}]{Lefeng Zhang} received his B.Eng. and M.Eng. degree from Zhongnan
University of Economics and Law, China, in 2016 and 2019, respectively, and a PhD degree from the University of Technology Sydney, Australia, in 2024. He is currently an assistant professor at the City University of Macau. His research interests are game theory and privacy-preserving.
\end{IEEEbiography}
\vspace{-3em}

\begin{IEEEbiography}[{\includegraphics[width=1in,height=1.25in,clip,keepaspectratio]{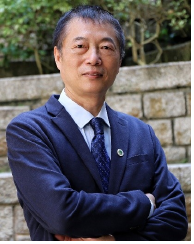}}]{Wanlei Zhou} (Senior member, IEEE) is currently the Vice Rector (Academic Affairs) and Dean of faculty of Data Science, City University of Macau, Macao SAR, China. He received the B.Eng and M.Eng degrees from Harbin Institute of Technology, Harbin, China in 1982 and 1984, respectively, and the PhD degree from The Australian National University, Canberra, Australia, in 1991, all in Computer Science and Engineering. He also received a DSc degree (a higher Doctorate degree) from Deakin University in 2002. Before joining City University of Macau, Professor Zhou held various positions including the Head of School of Computer Science in University of Technology Sydney, Australia, the Alfred Deakin Professor, Chair of Information Technology, Associate Dean, and Head of School of Information Technology in Deakin University, Australia. Professor Zhou also served as a lecturer in University of Electronic Science and Technology of China, a system programmer in HP at Massachusetts, USA; a lecturer in Monash University, Melbourne, Australia; and a lecturer in National University of Singapore, Singapore. His main research interests include security, privacy, and distributed computing. Professor Zhou has published more than 400 papers in refereed international journals and refereed international conferences proceedings, including many articles in IEEE transactions and journals. 
\end{IEEEbiography}
\vspace{-3em}
\begin{IEEEbiography}[{\includegraphics[width=1in,height=1.25in,clip,keepaspectratio]{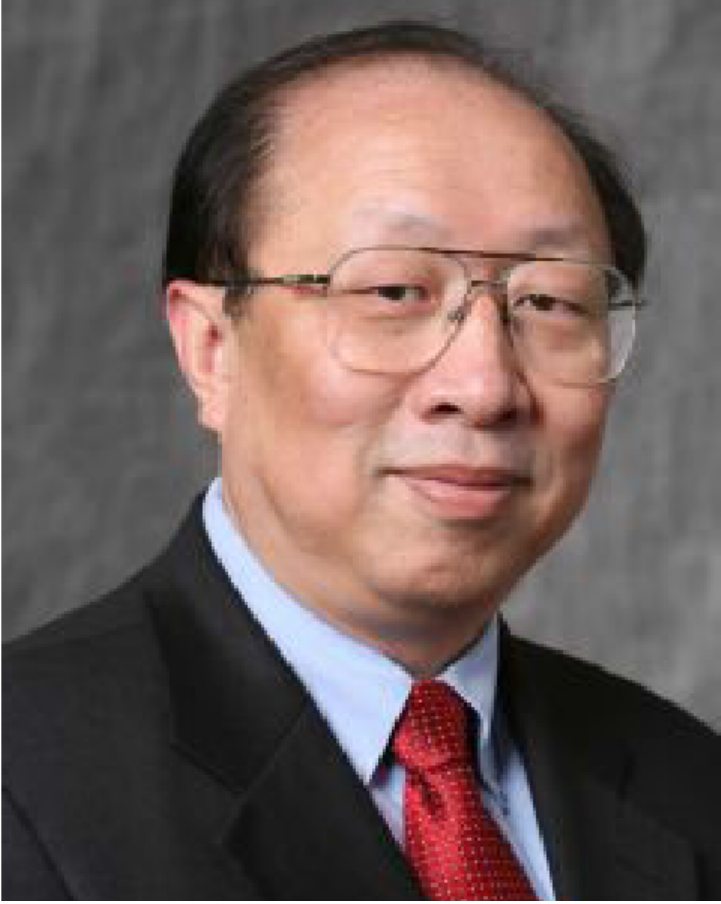}}]{Wei Zhao} (Fellow, IEEE) is currently the Chair Professor of Computer Science at Shenzhen Institute of Advanced Technology, University of Chinese Academy of Sciences, Shenzhen, China. He received a degree in physics from Shaanxi Normal University, China, in 1977 and the M.Sc. and Ph.D. degrees in computer and information sciences from the University of Massachusetts at Amherst in 1983 and 1986, respectively. He has served in important leadership roles in academics, including the Chief Research Officer with the American University of Sharjah; the Chair of the Academic Council; CAS Shenzhen Institute of Advanced Technology; the eighth rector of the University of Macau; the Dean of Science with the Rensselaer Polytechnic Institute; the Director for the Division of Computer and Network Systems, U.S. National Science Foundation; and the Senior Associate Vice President for Research at Texas A\&M University. He has made significant contributions to cyber-physical systems, distributed computing, real-time systems, and computer networks. He led the effort to define the research agenda and create the very first funding program for cyber-physical systems in 2006. His research results have been adopted as the standard for survivable, adaptable fiber optic embedded networks. He was awarded the Lifelong Achievement Award by the Chinese Association of Science and Technology in 2005.
\end{IEEEbiography}

\end{document}